\documentclass[preprint,12pt]{elsarticle}

\usepackage{amssymb}
\usepackage{amsmath}
\usepackage{amsthm}
\usepackage{hyperref}

\usepackage{subcaption}
\usepackage{booktabs}
\usepackage{color,soul}

\theoremstyle{plain}
\newtheorem{theorem}{Theorem}[section]
\newtheorem{proposition}[theorem]{Proposition}
\newcommand{\dv}{\boldsymbol{x}}
\newcommand{\dd}{\mathcal{X}}
\newcommand{\ds}{\boldsymbol{X}}
\newcommand{\pv}{\boldsymbol{c}}
\newcommand{\ps}{\boldsymbol{C}}
\DeclareMathOperator{\sech}{sech}

\journal{Information Sciences}

\begin{document}

\begin{frontmatter}

\title{Quantifying Behavioral Dissimilarity Between Mathematical Expressions}

\author[IJS,MPS,FMF]{Sebastian Me\v{z}nar}
\author[IJS]{Sa\v{s}o D\v{z}eroski}
\author[FMF,IJS]{Ljup\v{c}o Todorovski}

\affiliation[IJS]{organization={Jožef Stefan Institute},
            addressline={Jamova cesta 39}, 
            city={Ljubljana},
            postcode={1000}, 
            country={Slovenia}}

\affiliation[MPS]{organization={Jožef Stefan International Postgraduate School},
            addressline={Jamova cesta 39}, 
            city={Ljubljana},
            postcode={1000}, 
            country={Slovenia}}

\affiliation[FMF]{organization={Faculty of Mathematics and Physics, University of Ljubljana},
            addressline={Jadranska 21}, 
            city={Ljubljana},
            postcode={1000}, 
            country={Slovenia}}

\begin{abstract}
Quantifying the similarity between mathematical expressions is a fundamental problem in computational mathematics, symbolic reasoning, and scientific discovery. While behavioral notions of similarity have previously been explored in the context of software and program analysis, existing measures for mathematical expressions rely primarily on syntactic form, assessing similarity through symbolic structure rather than actual behavior. Yet syntactically distinct expressions can exhibit nearly identical outputs, while structurally similar ones may behave very differently---especially when the expressions contain free parameters that define families of functions. To address these limitations, we introduce Behavior-aware Expression Dissimilarity (BED), a principled framework for quantifying behavioral distance between mathematical expressions with free parameters. BED represents expressions as joint probability distributions over their input–output pairs and applies the Wasserstein distance to measure behavioral dissimilarity. A computationally efficient stochastic approximation is proposed and shown to be consistent, robust, and capable of inducing a smoother, more meaningful structure over the space of expressions than syntax-based measures. The approach provides a foundation for behavior-based comparison, clustering, and learning of mathematical expressions, with potential direct applications in equation discovery, symbolic regression, and neuro-symbolic modeling.
\end{abstract}


%
%
%
%


\end{frontmatter}


\setcounter{figure}{0}

\section{Introduction}
Measuring how mathematical expressions differ in their behavior---rather than in their syntax---is a fundamental problem in computational mathematics, machine learning, and symbolic reasoning. Understanding whether two expressions yield similar outputs across a range of inputs is essential for tasks such as model comparison, expression clustering, and the discovery of interpretable functional relationships. Despite its importance, most existing similarity measures are syntactic in nature, comparing expressions based on their structural form rather than their actual functional behavior. However, syntactically distinct expressions can often produce nearly identical outputs over a range of inputs, while minor syntactic changes can yield drastically different behaviors. This discrepancy motivates the development of quantitative measures that capture behavioral dissimilarity.

The concept of behavioral similarity has been studied in the context of program and code analysis, where program fragments are compared based on their execution behavior rather than their textual structure. For example, \citet{Walenstein2007DagSemProc, Juergens2010BeyondCopy} introduced behavioral notions of program similarity, demonstrating that behavioral measures can reveal equivalences between structurally diverse implementations. However, these approaches have not been extended to mathematical expressions, where additional challenges arise due to the presence of free parameters that define entire families of functions. Quantifying behavioral dissimilarity in this setting requires accounting for the variability in outputs induced by parameter changes, which demands a probabilistic treatment of expression behavior.

Existing approaches to tasks such as symbolic regression or equation discovery also implicitly rely on some notion of expression similarity when exploring the space of candidate equations. Traditionally, this search space is organized according to syntactic similarity. Evolutionary algorithms generate new expressions through syntactic transformations such as crossover and mutation~\cite{cranmer2023pysr}, while neural approaches learn latent representations in which nearby vectors correspond to syntactically similar expressions~\cite{Mežnar2023HVAE}. However, syntactic proximity offers only limited guidance for behavioral similarity, and this bias results in inefficient exploration, restricting the generalizability of learned models. Organizing the search space according to behavioral similarity has the potential to enhance exploration efficiency and yield representations that are more consistent with functional semantics.

To address these limitations, we introduce Behavior-aware Expression Dissimilarity (BED), a principled framework for quantifying behavioral distance between mathematical expressions with free parameters. BED represents expressions as joint probability distributions over their input and output spaces, thereby capturing behavioral variability induced by the parameters. Behavioral dissimilarity is then formalized using the Wasserstein distance, a robust metric for comparing probability distributions. Because BED cannot be computed analytically, we propose a stochastic approximation that is computationally feasible, consistent, and robust across diverse experimental settings.

We empirically evaluate the proposed BED approximation in two complementary directions. First, we assess its effectiveness in clustering behaviorally equivalent expressions. We hypothesize that expressions from the same class of functions (e.g., polynomial, trigonometric, or logarithmic) exhibit dissimilarity values close to zero, while non-equivalent ones from distinct classes are substantially more dissimilar. To test this hypothesis, we generate sets of equivalent expressions, compute pairwise dissimilarities using BED and syntax-based distance measures, cluster the resulting distance matrices, and evaluate the quality of the resulting clusters.  Second, we examine whether BED induces a smooth error manifold over the space of mathematical expressions with free parameters. This property is critical for symbolic regression and related search tasks: expressions that are close under BED should yield similar predictive errors on a given dataset. In such a landscape, decreasing dissimilarity to the ground truth should correspond to a reduction in error, thereby enabling more efficient and interpretable search. The central claim of this work is that BED substantially increases the smoothness of this error landscape compared to syntax-based alternatives.

Beyond its empirical validation, BED establishes a general foundation for reasoning about mathematical expressions in terms of their behavioral properties rather than their syntactic form. It enables behavior-based comparison, clustering, and learning of mathematical expressions from data, offering practical benefits for symbolic regression, equation discovery, and neuro-symbolic learning, as well as potential extensions to other parameterized computational models.

The remainder of the paper is organized as follows. Section~\ref{sec:background} establishes the basic terminology and introduces the notions of syntactic and behavioral similarity between expressions and, most notably, formalizes the behavioral representation of mathematical expressions as joint probability distributions over inputs and outputs. Section~\ref{sec:bed} defines the BED measure based on the Wasserstein distance and presents its stochastic approximation. In Section~\ref{sec:evaluation}, we first examine the consistency and robustness of the approximation, and then evaluate its performance for expression clustering and structuring the space of candidate expressions in symbolic regression. Section~\ref{sec:discussion} discusses the strengths and weaknesses of BED. Finally, Section~\ref{sec:conclusion} summarizes our findings and outlines directions for future work.

\section{Background and related work}\label{sec:background}
In this section, we delve into the definition of mathematical expressions and their behavior, particularly in the context of symbolic regression. We begin by formally defining mathematical expressions in Section~\ref{sec:expressions}, clarifying the roles of variables and constants, and introducing the concept of expressions with free parameters. Following this, we discuss the behavior of expressions in Section~\ref{sec:behavior}, focusing on how the behavior of expressions, especially those with free parameters, can be quantified as a probability distribution over their outputs. Next, we explore current measures for quantifying the dissimilarity between expressions in Section~\ref{sec:dissimilarity}. This section distinguishes between syntactic and behavioral approaches, highlighting their strengths and weaknesses. Finally, we conclude this section by highlighting the role of dissimilarity measures in symbolic regression in Section~\ref{sec:dmsr}.

\subsection{Mathematical expressions}\label{sec:expressions}
Mathematical expressions, typically written as a sequence of symbols in infix notation, offer a concise and interpretable way to model the relationship between input and output variables. For instance, the expression $2x+5$ models the relationship between the input $x$ and the output $y$ in the equation $y=2x+5$. These expressions are composed of unary and binary operators/functions, variables, and constants. Furthermore, expressions in infix notation often utilize parentheses to denote the order of operations, specifying which subexpressions should be calculated first.

Variables and constants both represent numeric values, but they play fundamentally different roles in an expression. A variable, such as $x$, represents an element from a domain and can have a different value at each data point, enabling expressions to model varying inputs. Unlike variables, constants retain fixed values regardless of the input and evaluate to the same value across all data points. For example, in the expression $2x+5$, $x$ is a variable, while $2$ and $5$ are constants.

In symbolic regression, we typically work with a more general form of expressions that contain free parameters, where fixed numerical constants are replaced by placeholder symbols such as $C_i$ (where $i$ is an index) or a generic symbol $C$. The distinction between these parameters and variables is crucial. Variables, such as $x$ in the expression $C_0 \cdot x + C_1$, are independent of the expression and are provided by the problem's data, which defines the domain of the expression. In contrast, free parameters like $C_0$ and $C_1$ are intrinsic to the expression's structure; their values are learned or optimized to fit the observed data. This use of free parameters significantly reduces the search space of possible expressions, as it eliminates the need to exhaustively search for specific parameter values. Consequently, the primary focus of symbolic regression shifts to identifying the correct structural form of the equation, after which the free parameters can be efficiently optimized to yield an accurate model.

\subsection{Behavior of expressions}\label{sec:behavior}
While mathematical expressions are conveniently represented using abstract notation for their conciseness and interpretability, our primary interest often lies in the functional mapping they describe. This mapping, which dictates the values of the output variable for every combination of input variables, effectively describes the behavior of a mathematical expression. For example, the expression $2 x + 5$ corresponds to the mapping $x \mapsto 2 x + 5$. Throughout this paper, we will refer to this functional mapping, or the curve/manifold it produces, as the expression's behavior.

An expression with fixed constant values exhibits deterministic behavior, meaning it produces a unique output for every combination of input variables within its domain. In contrast, when considering mathematical expressions with free parameters, each combination of input variables corresponds to a set of outputs that can vary based on the particular values assigned to these parameters. Since values from this set of outputs can occur multiple times (e.g., if different parameter values lead to the same output for a given input), we can define a probability distribution over them. In this way, the behavior of a given expression with free parameters can be viewed and analyzed as a probability distribution.

To define this probability distribution, we first consider a single input point $\dv \in \dd$ and an expression $\mathit{E}$ with $p$ free parameters. The values for these parameters are stored in a vector $\pv = (c_1, c_2, \dots, c_p)$ from the domain $\mathcal{C}^p$. We write the output of $\mathit{E}$ for a given input $\dv$ and parameter vector $\pv$ as $\mathit{E}(\dv, \pv)$. The set of all possible outputs of $\mathit{E}$ at $\dv$, which forms the support of our probability distribution, is then denoted as $\mathit{E}(\dv, \mathcal{C}^p) := \{\mathit{E}(\dv, \pv) \mid \pv \in \mathcal{C}^p\}$.

If we assume that the values of the free parameters $\pv$ are drawn from a probability distribution, say $P(\pv)$ defined over their domain $\mathcal{C}^p$ (e.g., a uniform distribution over the hypercube $[-5,5]^p$), then for a fixed input point $\dv$, the output $\mathit{E}(\dv, \pv)$ becomes a random variable. The behavior of expression $\mathit{E}$ at the point $\dv$ is then fully described by the probability distribution of this random variable.

\begin{figure*}[h!]
  \centering
  \includegraphics[width=\linewidth]{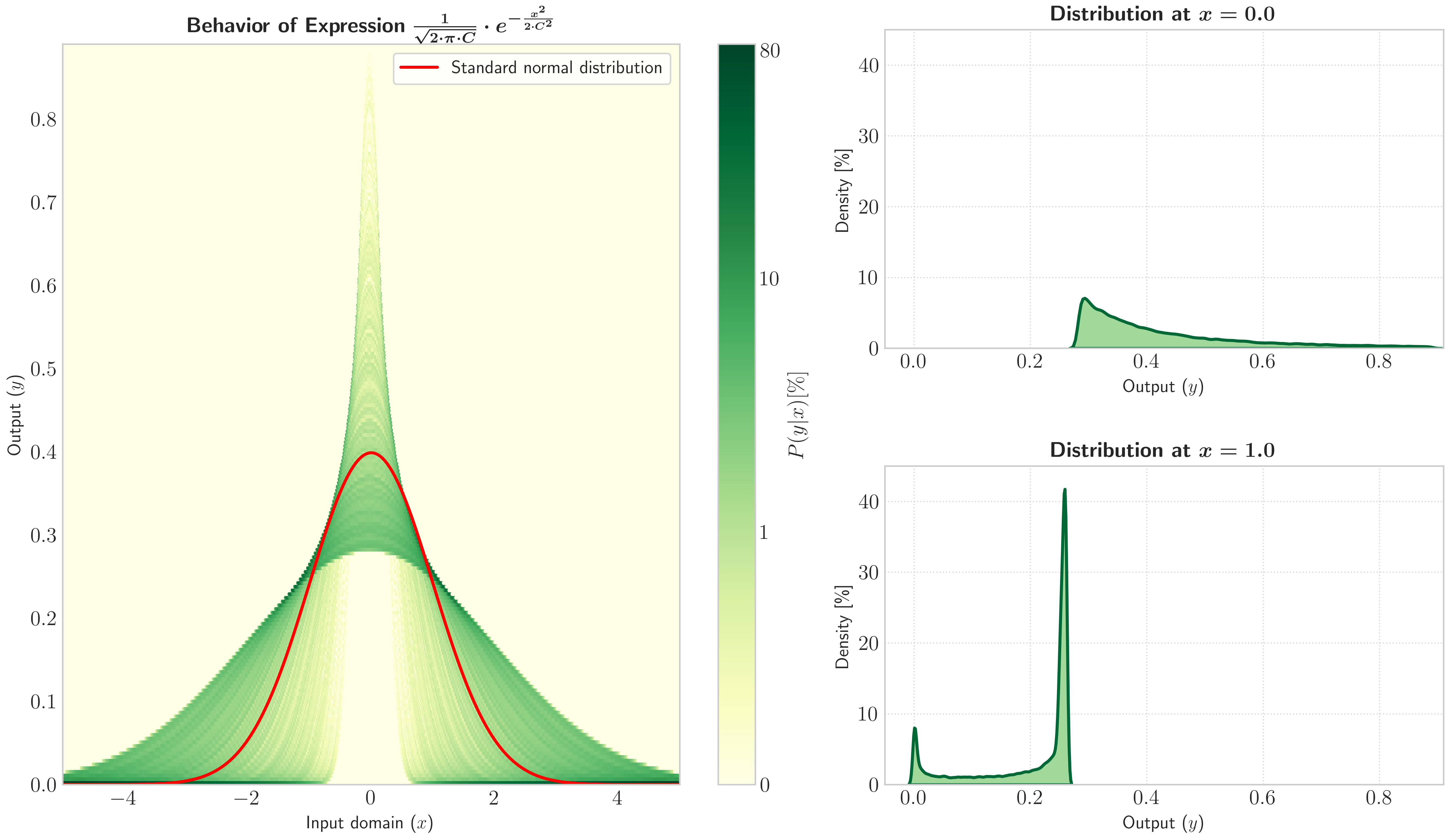}
  \caption{The left-hand side shows the behavior of the expression $y = \frac{1}{\sqrt{2\pi C}} \cdot e^{\frac{-x^2}{2 \cdot C^2}}$ modeled as a probability distribution. This figure illustrates how we define the behavior of an expression with free parameters. The distribution is obtained by sampling the free parameter $C$ (representing standard deviation) from the interval $[0.2, 2]$. The color intensity at each point indicates the conditional probability density of an output value $y$ given an input $x$. A higher density (green color) means that more sampled curves pass through that point. The red line shows the standard normal distribution, representing the behavior of a single expression with a fixed parameter value of $C = 1$. The right-hand side shows the distributions of output of the expression in the left figure at $x=0$ and $x=1$.}
  \label{fig:behavior}
\end{figure*}

Extending this to the entire input domain $\dd$, the overall behavior of an expression $\mathit{E}$ with free parameters can be defined as a joint probability distribution over its input and output values. Specifically, by considering inputs $\dv$ as random variables over $\dd$ (e.g., uniformly distributed) and parameters $\pv$ as random variables drawn from their distribution $P(\pv)$ over $\mathcal{C}^p$, the expression $\mathit{E}$ implicitly defines a joint distribution of the form $P(\dv, \boldsymbol{y})$, where $\boldsymbol{y} = \mathit{E}(\dv, \pv)$. This distribution effectively captures the likelihood of observing specific output values for any given input, taking into account the variability introduced by the free parameters. Figure~\ref{fig:behavior} shows an example of behavior, specifically the behavior of the expression for the density of the normal distribution, where we view standard deviation as a free parameter.

\subsection{Dissimilarity between mathematical expressions}\label{sec:dissimilarity}
Quantifying the dissimilarity between objects is a fundamental task in many fields. A dissimilarity measure is a non-negative, symmetric function that intuitively captures how different two objects are. However, it is a relaxation of a formal distance metric. A distance metric must also satisfy two key axioms: the identity of indiscernibles ($d(x,y)=0 \Leftrightarrow x=y$) and the triangle inequality ($d(x,z)\leq d(x,y)+d(y,z)$).

The choice of a dissimilarity measure between mathematical expressions depends on the specific task. For instance, tasks like searching for structurally similar expressions within a database might require a syntax-based measure, which views expressions as a sequence of symbols. For other tasks, such as function approximation or symbolic regression, a behavioral-based dissimilarity measure is more suitable, as it focuses on the functional mapping that the expression defines. We will use the term behavioral-based dissimilarity measure consistently throughout this work because, in the context of symbolic regression and other scientific disciplines, 'semantics' are often associated with a higher-level understanding or interpretation of an expression—for example, interpreting the equation $v=a\cdot t$ as "the formula for velocity," rather than merely referring to the manifold produced by the expression $a\cdot t$.

\subsubsection{Syntactic dissimilarity}
Syntactic dissimilarity measures the difference between two mathematical expressions based on their structure, treating them as sequences of symbols or hierarchical trees. Some examples of syntactic dissimilarity measures include the Levenshtein distance~\cite{Levenshtein1965BinaryCC} (also known as edit distance), tree edit distance~\cite{Zhang1989Treeedit}, Jaro~\cite{Jaro01061989}, Longest Common Subsequence (LCS), and Jaccard distance. Edit distance quantifies the dissimilarity between expressions' string representations (e.g., in the infix notation), while tree edit distance measures the difference between their corresponding tree structures. Both calculate dissimilarity as the minimal number of elementary edit operations (such as adding, removing, or editing a symbol or a tree node) required to transform one expression into another. These dissimilarity measures operate directly on the symbolic representations of mathematical expressions and are thus in the context of code similarity~\cite{Walenstein2007DagSemProc} sometimes considered instances of \emph{representational} similarity.

By design, these syntax-based dissimilarity measures are well-suited for mathematical expressions containing free parameters. Since free parameters are typically represented as a distinct token (e.g., $C_0$, $C_1$, or $C$), they are treated like any other symbol during the dissimilarity calculation. This characteristic allows the dissimilarity between expressions like $2x+5$ and $2x-3$ to be $1$ (a single edit to change the $+$ symbol into $-$), same as the dissimilarity between $2x+5$ and $2x-5$. A further advantage of these dissimilarities is their utility in generating new, similar expressions; one can simply modify a token or node while ensuring syntactic correctness. In the context of symbolic regression, this aspect implicitly promotes the parsimony principle, as longer or more complex expressions inherently have higher dissimilarity to the initial, usually shorter expressions.

However, syntactic dissimilarity measures don't directly correlate with behavioral similarity. While this correlation might hold for general trends or broad structural comparisons, it often breaks down in more critical scenarios. For instance, two expressions can be syntactically very different yet behave identically (e.g., $x+x+x+x$ vs.~$4 x$), or conversely, they can be syntactically nearly identical but exhibit vastly different behaviors (e.g., $x^2$ vs.~$x^5$ or $\sin x$ vs.~$\cos x$ for small changes in input). This divergence means that a syntactically minor change can lead to a drastic shift in an expression's output, posing a significant limitation when the goal is to find expressions that closely match a target behavior.

\subsubsection{Behavioral dissimilarity}
Behavioral dissimilarity measures how different two mathematical expressions are based on their behavior across a defined domain of input variables. This approach directly compares the numerical values or probability distributions they produce, irrespective of their symbolic representation. Unlike syntactic dissimilarity, which is indifferent to the domain of input variables, behavioral dissimilarity critically depends on this defined input domain, as an expression's output—and thus its behavior—can vary significantly across different domains. We proceed by assuming a known input domain, with a discussion on the influence of domain selection deferred to Section~\ref{sec:discussion}.

Despite the importance of behavioral comparison, research into dedicated behavioral dissimilarity measures for mathematical expressions, particularly those containing free parameters, remains relatively limited. However, for expressions exhibiting deterministic behavior (i.e., those with fixed constant values) or, more commonly, for quantifying the dissimilarity between an expression's behavior and ground truth data, established statistical measures can be adapted. The most used measures for this purpose include the Root Mean Squared Error (RMSE) and the Coefficient of Determination ($R^2$). 

RMSE is defined as $\text{RMSE}(\boldsymbol{y}, \hat{\boldsymbol{y}})=\sqrt{\frac{1}{n}\sum_{i=1}^n (\boldsymbol{y}_i - \hat{\boldsymbol{y}_i})^2}$, where $n$ is the number of data points, $\hat{y_i}$ is the output of an expression at input $\dv$, and $\boldsymbol{y}_i$ is either the output of another expression at input $\dv$ or the ground truth corresponding to this input. Because of its symmetric nature, RMSE can be used to measure the behavioral dissimilarity between two expressions with fixed constant values or the dissimilarity of an expression with fixed constant values to the ground truth data.

On the other hand, the $R^2$ metric is asymmetric, and thus can only be used to assess how well an expression's outputs explain the variance in the ground truth data. It is defined as:

$$R^2(\boldsymbol{y}, \hat{\boldsymbol{y}}) = 1 - \frac{\sum_{i=1}^n (\boldsymbol{y}_i - \hat{\boldsymbol{y}}_i)^2}{\sum_{i=1}^n (\boldsymbol{y}_i - \bar{\boldsymbol{y}})^2} .$$
Here, $\bar{\boldsymbol{y}}$ represents the mean of the ground truth data.

While these behavioral dissimilarity measures are easy to compute, sensible, and work well with expressions exhibiting deterministic behavior or when comparing an expression against ground truth, they cannot be directly adapted to expressions with free parameters. This is because RMSE and $R^2$ are designed to compare sequences of output values, not probability distributions of outputs that arise from varying free parameters. Consequently, a significant gap remains in developing dedicated behavioral dissimilarity measures that can robustly compare these probabilistic behaviors.

\subsection{Dissimilarity measures in symbolic regression}\label{sec:dmsr}
A key application area for dissimilarity measures between mathematical expressions is symbolic regression. In this field, dissimilarity measures are predominantly used indirectly for structuring the search space by guiding the generation of new expressions, rather than solely for quantifying the dissimilarity between existing ones. Consequently, most—if not all—current approaches rely on syntactic dissimilarity measures in some way.

Consider genetic programming (GP)~\cite{Koza1994, Schmidt2009DistillingNaturalLaws, cranmer2023pysr}, a prevalent approach in symbolic regression. GP typically represents mathematical expressions as expression trees, and its core operations, like crossover and mutation, involve sequences of elementary edits to these trees. These operations inherently align with the concept of syntactic tree edit distance. Similarly, grammar-based methods~\cite{Todorovski1997BiasED,Brence2021ProGED} explore the space of parse trees, often prioritizing simpler structures and systematically generating more complex ones through minor revisions based on grammar rules. This ensures that expressions close in the search space are syntactically similar. Analogously, process-based modeling~\cite{Bridewell2008InductiveProcessModeling} systematically explores candidate models by enumerating expressions via simple edit operations, such as adding or changing processes.

In recent years, a growing body of research has explored the use of generative neural networks to embed expressions into a latent space where expressions with similar syntactic structures occupy close proximity~\cite{Kusner2017GVAE,Gomez2018CVAE,Mežnar2023HVAE}. These approaches then employ optimization algorithms to navigate this latent space and generate new expressions with similar syntactic properties. While these approaches do not explicitly use edit distance or tree edit distance during training, the underlying structure of the latent space aligns with these measures, effectively clustering syntactically similar expressions together. DSO~\cite{Petersen2021CharGen} further extends this concept by incorporating reinforcement learning and evolutionary algorithms to guide the search for new expressions. However, this approach's spatial relationships between expressions still largely mirror the neighborhoods defined by edit and tree edit distance metrics. Finally, Symbolic-Numeric Integrated Pre-training (SNIP)~\cite{meidani2023snip} diverges from this trend by employing two transformer-based encoders, one for symbolic and one for numeric data. Note, however, that SNIP requires a subsequent fine-tuning stage that utilizes syntactic similarity measures.

\section{Behavioral dissimilarity between expressions with free parameters}\label{sec:bed}
The main purpose of this paper is the introduction of a behavioral dissimilarity measure specifically designed for mathematical expressions containing free parameters. We first present the theoretical formulation of this proposed measure in Section~\ref{sec:definition}. Recognizing that this theoretical version is computationally intractable, we subsequently introduce its approximate version in Section~\ref{sec:approximation} to enable practical application.

\subsection{Behavior-aware expression dissimilarity (BED)}\label{sec:definition}
In Section~\ref{sec:behavior}, we present the behavior of expressions with free parameters as a probability distribution over their output values, arising from the variability in the parameter values. Consequently, dissimilarity measures from probability theory become particularly relevant. One promising choice is the $p$-Wasserstein distance, also known as Earth Mover's Distance~\cite{Villani2009}. The Wasserstein distance measures the minimum "work" required to transform one distribution into another. This metric is exceptionally well-suited for our problem because it inherently considers the differences between entire distributions, unlike simpler measures that merely compare means or variances. Furthermore, it has been shown to effectively capture intuitive notions of similarity, including human perception in certain contexts~\cite{Rubner2000}.

The $p$-Wasserstein distance is defined by Eq.~\eqref{eq:emd}, where $\Gamma(U, V)$ represents the set of all couplings between random variables $U$ and $V$, i.e., pairs of random variables ($u$, $v$), such that $u$ and $v$ have the same distributions as $U$ and $V$, respectively. Through the rest of the paper, we will focus on the $1$-Wasserstein distance.

\begin{equation}\label{eq:emd}
    W_p(U, V) = \bigg(\inf_{\pi \in \Gamma(U,V)}\int_{\mathbb{R}\times\mathbb{R}} |u-v|^p \, d\pi(u,v) \bigg)^\frac{1}{p}
\end{equation}

If we assume that the expressions evaluate to real numbers (i.e., their outputs are one-dimensional), the calculation of the Wasserstein distance can be significantly simplified. In this specific case, the distance $d(u,v)$ between individual output values $u$ and $v$ is naturally taken as $|u-v|$. With this assumption, the Wasserstein distance can be directly computed using the quantile function (or inverse cumulative distribution function) of the corresponding random variable. If $Q_{U}(q)$ and $Q_{V}(q)$ denote the quantile functions for distributions $U$ and $V$ respectively, the Wasserstein distance is given by Eq.~\eqref{eq:emd-r}~\cite{Ramdas2017Wasserstein}.

\begin{equation}\label{eq:emd-r}
    W_p(U, V) = \int_{0}^{1} \left|Q_U(q) - Q_V(q)\right| \, dq
\end{equation}

This simplification offers substantial computational advantages because, for one-dimensional real random variables, the optimal coupling is unique and directly determined by the quantile functions, thereby avoiding the complex search for the infimum over all possible couplings. Furthermore, this simplified equation provides a more intuitive interpretation of the distance, as it directly quantifies the differences between the shapes and locations of the distributions' quantile functions.

We can now employ the Wasserstein distance from Eq.~\eqref{eq:emd} to quantify the dissimilarity between two expressions, $\mathit{E}$ and $\mathit{F}$, at a specific input point in the domain, $\dv\in\dd$. By denoting the probability distribution of expression $\mathit{E}$'s output at $\dv$ as $P_{\mathit{E}}(\dv)$, and that of expression $\mathit{F}$'s output as $P_{\mathit{F}}(\dv)$, we define the behavior-aware expression dissimilarity (BED) between $\mathit{E}$ and $\mathit{F}$ at $\dv$ as:

\begin{equation}\label{eq:bedxi}
    \text{BED}^{(\dv)}(\mathit{E},\mathit{F}) = \int_{0}^{1} \left|Q_{P_{\mathit{E}}(\dv)}(q) - Q_{P_{\mathit{F}}(\dv)}(q)\right| \, dq .
\end{equation}
Here, it is crucial to consider two edge cases regarding expression definition. First, if both expressions $\mathit{E}$ and $\mathit{F}$ are undefined for a given input point $\dv$ (e.g., due to division by zero for all combinations of their free parameter values), it is reasonable to set their dissimilarity at this point to $0$. This policy treats mutual undefinedness as an equivalent outcome, indicating no difference in behavior. A more challenging edge case arises when one expression is defined at $\dv$ but the other is not. For this scenario, we propose setting the dissimilarity between the two expressions at such a point to $+\infty$, thereby indicating a maximal dissimilarity due to a fundamental difference in their domain.

Finally, we extend this point-wise comparison to define the dissimilarity between two expressions with free parameters across the entire input domain $\dd$. This is achieved by integrating the point-wise $\text{BED}^{(\dv)}(\mathit{E},\mathit{F})$ (from Eq.~\eqref{eq:bedxi}) over the domain and normalizing the result by the domain's volume. The formal definition of BED is given by Eq.~\eqref{eq:bed}, where $\text{Vol}(\dd)$ represents the volume of the input domain $\dd$.

\begin{equation}\label{eq:bed}
    \text{BED}^{(\dd)}(\mathit{E},\mathit{F}) = \frac{1}{\text{Vol}(\dd)} \int_{\dd} \text{BED}^{(\dv)}(\mathit{E},\mathit{F}) \, d\dd
\end{equation}

Note that when both expressions are deterministic (i.e., contain no free parameters), Eq.~\eqref{eq:bedxi} simplifies to the absolute difference between their outputs at $\dv$: $|\mathit{E}(\dv, \emptyset)-\mathit{F}(\dv, \emptyset)|$. Consequently, Eq.~\eqref{eq:bed} reduces to the Mean Absolute Error (MAE) evaluated across all points in $\dd$. This simplification highlights the connection between BED and established metrics; for instance, if a $2$-Wasserstein distance were employed in Eq.~\eqref{eq:bedxi}, our measure could analogously be simplified to the Root Mean Squared Error (RMSE).

In \ref{app:characteristics}, we delve into the mathematical properties of the BED measure. We demonstrate that BED is a metric, since it is non-negative, symmetric, the dissimilarity of an expression to itself is zero, and the triangle inequality holds. We also establish that the identity of indiscernibles holds when expressions are viewed as probability distributions, but not when they are considered as symbolic sequences. While BED is a metric, we call it a dissimilarity measure though the rest of the paper because the definition from this section is computationally intractable and the properties don't hold for the aproximated version used though the rest of the paper.

\subsection{Approximating BED}\label{sec:approximation}
While the Behavior-aware Expression Dissimilarity (BED) offers a theoretical framework for quantifying the behavior-based dissimilarity between expressions, its direct analytical computation is generally intractable. This intractability stems from several challenges: the difficulty in deriving closed-form analytical expressions for the output probability distributions of any given expression with free parameters; the subsequent inability to obtain analytical quantile functions by inverting these complex distributions; and the difficulty of analytically solving the nested integrals involved in both the point-wise and overall BED definitions. To address these computational challenges, we employ an approximation strategy, which is outlined in detail in the remainder of this section.

\begin{figure*}[h!]
  \centering
  \includegraphics[width=\linewidth]{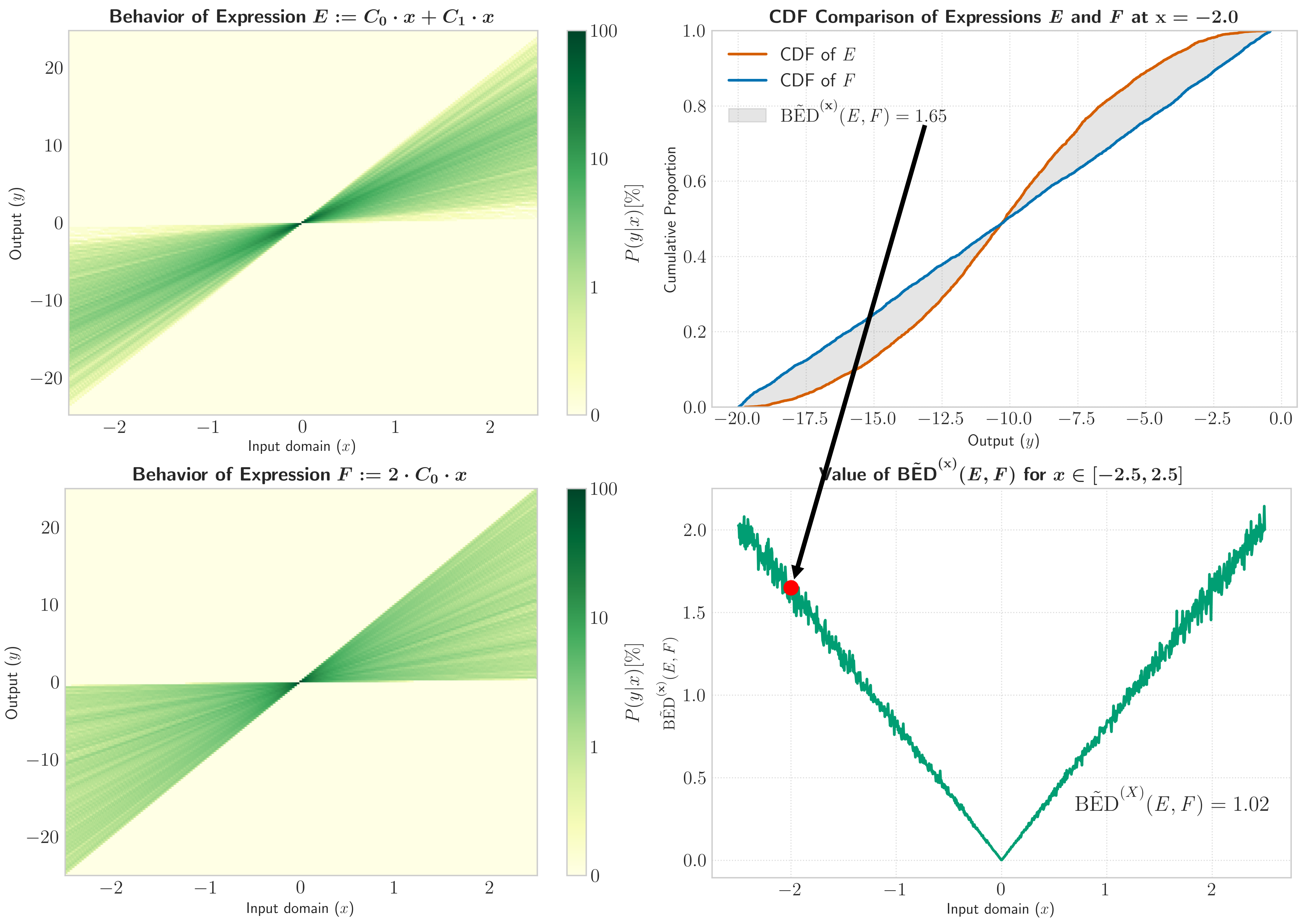}
  \caption{A visual breakdown of BED measure calculation. \textbf{Left panels} show the behaviors of expressions, $E := C_0 \cdot x + C_1 \cdot x$ and $F := 2 \cdot C_0 \cdot x$, across the domain $x \in [-2.5, 2.5]$. Values of the free parameters are sampled from the interval $[0.1, 5]$. The color bar on the right of each plot represents the probability of a given output value. \textbf{Top-right panel} compares the Cumulative Distribution Functions (CDFs) of both expressions at a specific point, $x=-2.0$. The shaded area between the two curves represents a single-point BED value between the expressions. \textbf{Bottom-right panel} plots the single-point BED values across the entire domain. The final BED score is the average of these values.}
  \label{fig:bed_example}
\end{figure*}

First, we address the approximation of the integral over the input domain $\dd$ in Eq.~\eqref{eq:bed}. This is achieved by discretizing the input domain, replacing the continuous integration with a summation over a finite set of sampled input points. This set of $n$ points, which we denote as $\ds=\{\dv_1, \dots, \dv_n\}$, can be formed in two ways: either by utilizing an existing dataset relevant to the application, or by assigning a probability distribution over the input domain $\dd$ and sampling from it. While different BED evaluations may utilize different input sets, it is crucial that both expressions ($\mathit{E}$ and $\mathit{F}$) are evaluated on the same set of points $\textbf{X}$ when calculating the dissimilarity between them. In our implementation of BED, we employ Latin Hypercube Sampling (LHS)~\cite{McKay1979LHS} to ensure an efficient and representative sampling of the input domain.

Next, we approximate the behavior of the expressions whose dissimilarity we want to calculate. Given an expression $\mathit{E}$ with $p$ free parameters, we first sample a set of $m$ parameter vectors denoted by $\ps=\{\pv_1, \dots, \pv_m\}$, from the probability distribution $P(\pv)$ defined over the parameter domain $\mathcal{C}^p$. We then evaluate the expression $\mathit{E}$ for each input point $\dv$ from the sampled input set $\ds$ and for each sampled parameter vector $\pv_j\in\ps$. This process yields a matrix $Y^{(\textit{E})}$ of size $n\times m$, where elements represent the output for a given input point and set of parameter values, formalized as $Y^{(\textit{E})}_{i,j} = \mathit{E}(\dv_i, \pv_j)$. We treat this matrix $Y^{(\textit{E})}$ as an approximation of the behavior of expression $\mathit{E}$. Similar to input points, our implementation also employs Latin Hypercube Sampling for sampling vectors containing parameter values.

To approximate the BED measure for a specific input $\dv$, as defined by Eq.~\eqref{eq:bedxi}, we leverage the sampled behavior matrices from the previous step. For each input point, we extract the corresponding rows of the $Y^{(\textit{E})}$ and $Y^{(\textit{F})}$ matrices, which represent output samples for expressions $\mathit{E}$ and $\mathit{F}$. From these samples, we then construct an empirical cumulative distribution function (CDF) for each expression, defined as the proportion of its samples that are less than or equal to a given value $y$. This is formulated in Eq.~\eqref{eq:cdf}.

\begin{equation}\label{eq:cdf}
\Tilde{F}_{P_{\mathit{E}}(\dv)}(y) = \frac{1}{m} \sum_{s \in Y^{(\textit{E})}_{i,:}} \mathbf{1}_{\{s \leq y\}}
\end{equation}

To compute the point-wise BED, we first define a combined set of all unique output values from both sample sets, denoted as $S_i$. Let $y_k$ be the $k$-th value in this sorted set, and $d_k = y_{k+1} - y_k$ be the distance between consecutive values. The approximation of BED at $\dv$, $\Tilde{\text{BED}}^{(\dv)}(\mathit{E}, \mathit{F})$, is then given by Eq.~\eqref{eq:abedxi}. This formula numerically approximates the integral of the absolute difference between the empirical CDFs by performing a summation over the intervals defined by the combined set of outputs.

\begin{equation}\label{eq:abedxi}
    \Tilde{\text{BED}}^{(\dv)}(\mathit{E}, \mathit{F}) = \sum_{k=1}^{|S_i|-1} d_k\left| \Tilde{F}_{P_{\mathit{E}}(\dv)}(y_k)-\Tilde{F}_{P_{\mathit{F}}(\dv)}(y_k) \right|
\end{equation}

Finally, to obtain the overall approximation of BED, which corresponds to the definition in Eq.~\eqref{eq:bed}, we replace the continuous integral over the input domain X with a summation over our sampled input set X. Specifically, the overall approximated BED is calculated as the average of approximations of BED (as computed by Eq.~\eqref{eq:abedxi}) over all points in X. This approximation is formally defined in Eq.~\eqref{eq:abed}.

\begin{equation}\label{eq:abed}
    \Tilde{\text{BED}}^{(\textbf{X})}(\textit{E}, \textit{F}) = \frac{1}{|\mathcal{X}|}\sum_{\dv\in \mathcal{X}} \Tilde{\text{BED}}^{(\dv)}(\textit{E}, \textit{F})
\end{equation}

We will use this approximation of BED throughout the rest of this paper to quantify the dissimilarity between two expressions. Figure~\ref{fig:bed_example} provides a visual breakdown of our measure, where we compare expressions $E := C_0 \cdot x + C_1 \cdot x$ and $F := 2 \cdot C_0 \cdot x$. Though they look deceptively similar, they are not truly equivalent, as $E$ contains two independent parameters ($C_0, C_1$) while $F$ contains only one. This difference in parameter independence slightly alters the distributions of their outputs, similar to how the distribution of the sum of two independent dice rolls is different to the distribution of a single die roll multiplied by two.

\section{Evaluation}\label{sec:evaluation}
In this section, we report on the results of assessing the proposed measure's validity and effectiveness. First, we demonstrate the consistency of our stochastic distance measure with experiments in Section~\ref{sec:consistency}. Next, we evaluate the measure's ability to cluster behaviorally equivalent expressions in Section~\ref{sec:clustering}. Finally, we investigate the smoothness of the error manifold. By using our measure to order expressions relative to the ground truth, we assess whether proximity in the expression space reliably correlates with a reduction in error on a specific dataset. This crucial property, necessary for guiding symbolic search algorithms, is quantitatively confirmed in Section~\ref{sec:smoothness}.

In the experiments presented in Section~\ref{sec:consistency} and Section~\ref{sec:smoothness}, we generate random expressions by sampling a probabilistic context-free grammar for mathematical expressions presented in~\ref{app:grammar}. The code for reproducing the results of the evaluation experiments can be found at the following link: \href{https://github.com/smeznar/BED}{https://github.com/smeznar/BED}. Additionally, a streamlined implementation of BED can be found included in the \href{https://github.com/smeznar/SymbolicRegressionToolkit}{SymbolicRegressionToolkit Python package}.

\subsection{Impact of sampling on approximation}\label{sec:consistency}
A crucial property of any dissimilarity measure is its consistency. While deterministic measures produce identical results across repeated evaluations, the proposed BED measure involves stochastic components due to sampling. Therefore, it is necessary to verify that its outputs remain stable when the dissimilarity between pairs of expressions is computed multiple times with independently sampled inputs.

Because dissimilarity values may vary greatly in magnitude, directly measuring consistency using statistics such as the standard deviation can be misleading. Instead, we focus on the relative ordering of expressions induced by the dissimilarity measure. Specifically, we assess consistency by examining whether BED similarly ranks expressions across repeated runs. To this end, we compute the Spearman’s rank correlation coefficient~\cite{Spearman1904rho} between rankings obtained in independent stochastic evaluations.

The approximation quality and computational efficiency of BED are primarily influenced by two hyperparameters: the number of sampled input points $|\ds|$ and the number of sampled parameter vectors $|\mathbf{C}|$. We investigate their impact by evaluating consistency on a set of 200 randomly sampled expressions with at most two variables and an arbitrary number of free parameters; results for expressions with different numbers of variables are provided in \ref{app:consistency}.

For each hyperparameter configuration, we calculate the distance between each expression and every other expression and rank the expressions accordingly. This stochastic ranking procedure is repeated $100$ times, yielding $100$ rankings per expression. We then compute the Spearman’s rank correlation coefficient for all pairs of rankings corresponding to the same expression and average these values to obtain a consistency score for that expression. The final consistency measure is the average of these scores over all 200 expressions, where higher values indicate greater stability across runs.

To systematically explore the hyperparameter space, we consider all pairwise combinations of $|C|$ and $|\ds|$, each ranging over $\{4, 8, 16, 32, 64, 128\}$, resulting in 36 distinct configurations. Each configuration is evaluated independently using the procedure described above.

We additionally consider three strategies for sampling the input space. In the first, a single set of input points $\ds$ is sampled once and reused for all pairwise distance computations in all $100$ runs. In the second, a new set of input points is sampled for each of the $100$ runs. In the third, a new set of input points is sampled for each pair of expressions. In all strategies, input points and parameter vectors are sampled using Latin hypercube sampling (LHS) from a uniform distribution over $[1,5]$. This interval is selected to avoid regions where elementary functions used in the expressions (such as the square root and logarithm) are undefined or numerically unstable.

\begin{figure}[h!]
    \centering    
    \begin{subfigure}[b]{0.49\textwidth}
        \centering
        \includegraphics[width=\textwidth]{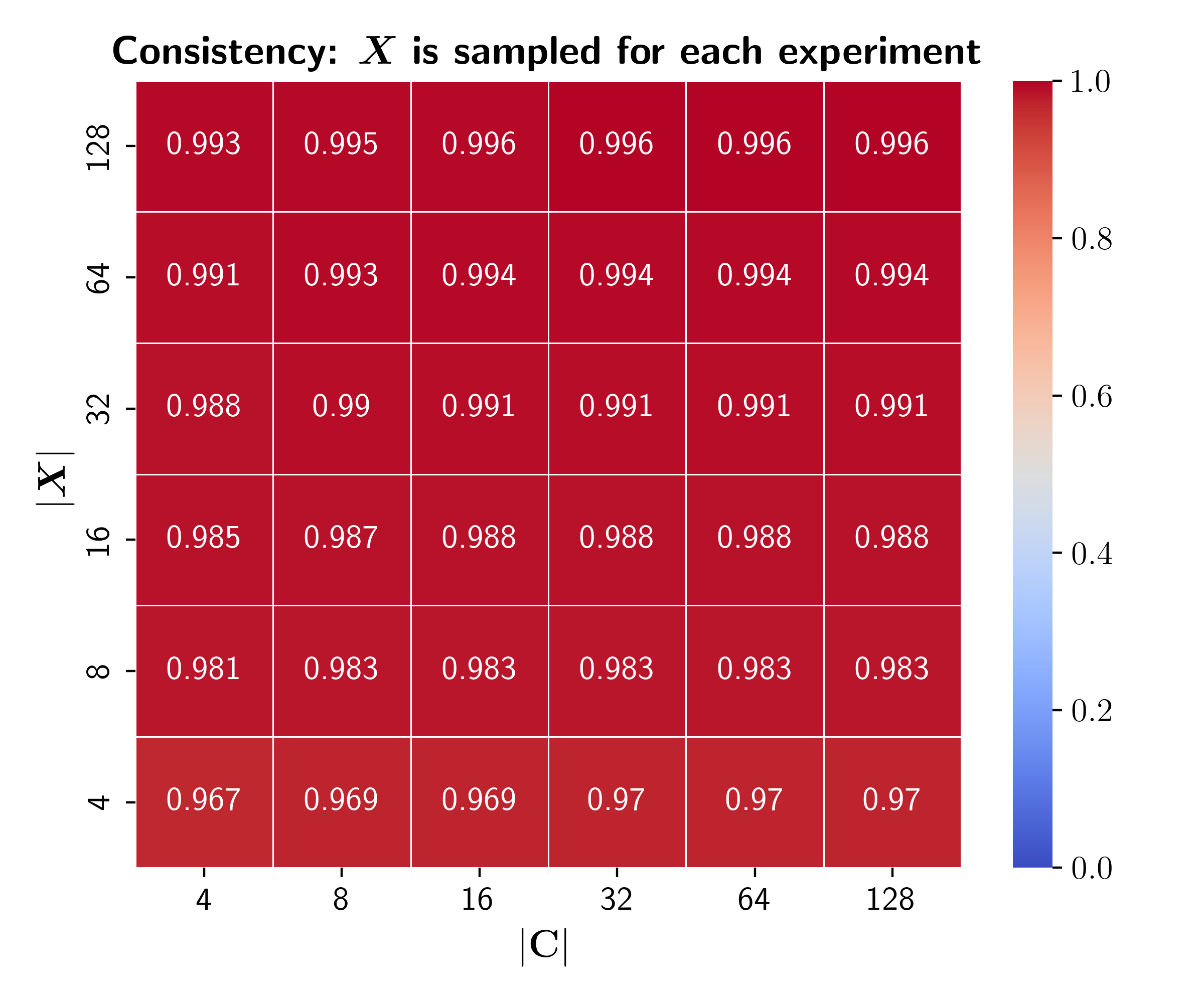} 
        \setcounter{subfigure}{1} 
        \caption{New $\ds$ sampled for each experiment.}
        \label{fig:consistency_run}
    \end{subfigure}
    
    \begin{subfigure}[b]{0.49\textwidth}
        \centering
        \includegraphics[width=\textwidth]{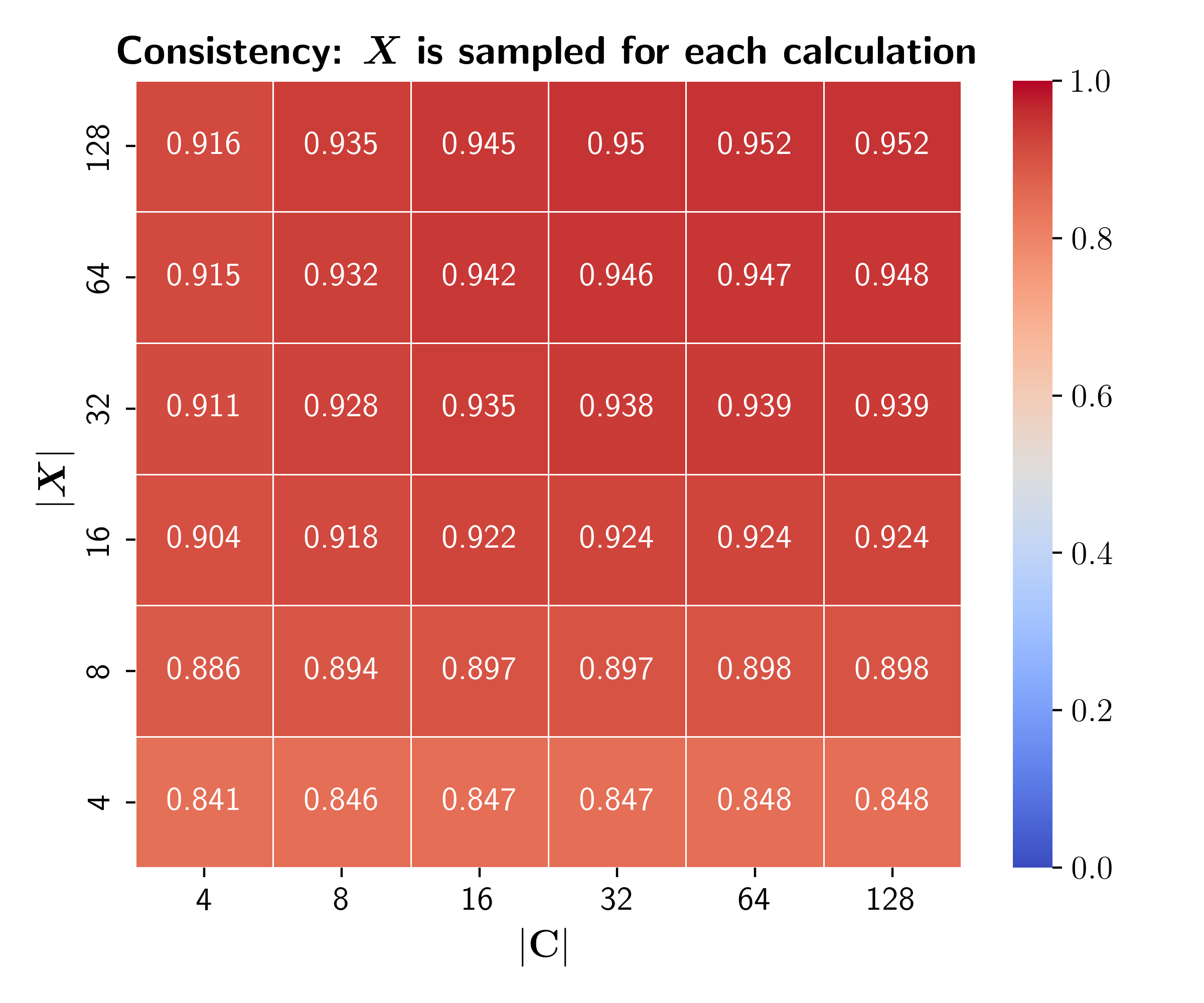} 
        \setcounter{subfigure}{0}
        \caption{New $\ds$ sampled for each calculation.}
        \label{fig:consistency_pair}
    \end{subfigure}
    \begin{subfigure}[b]{0.49\textwidth}
        \centering
        \includegraphics[width=\textwidth]{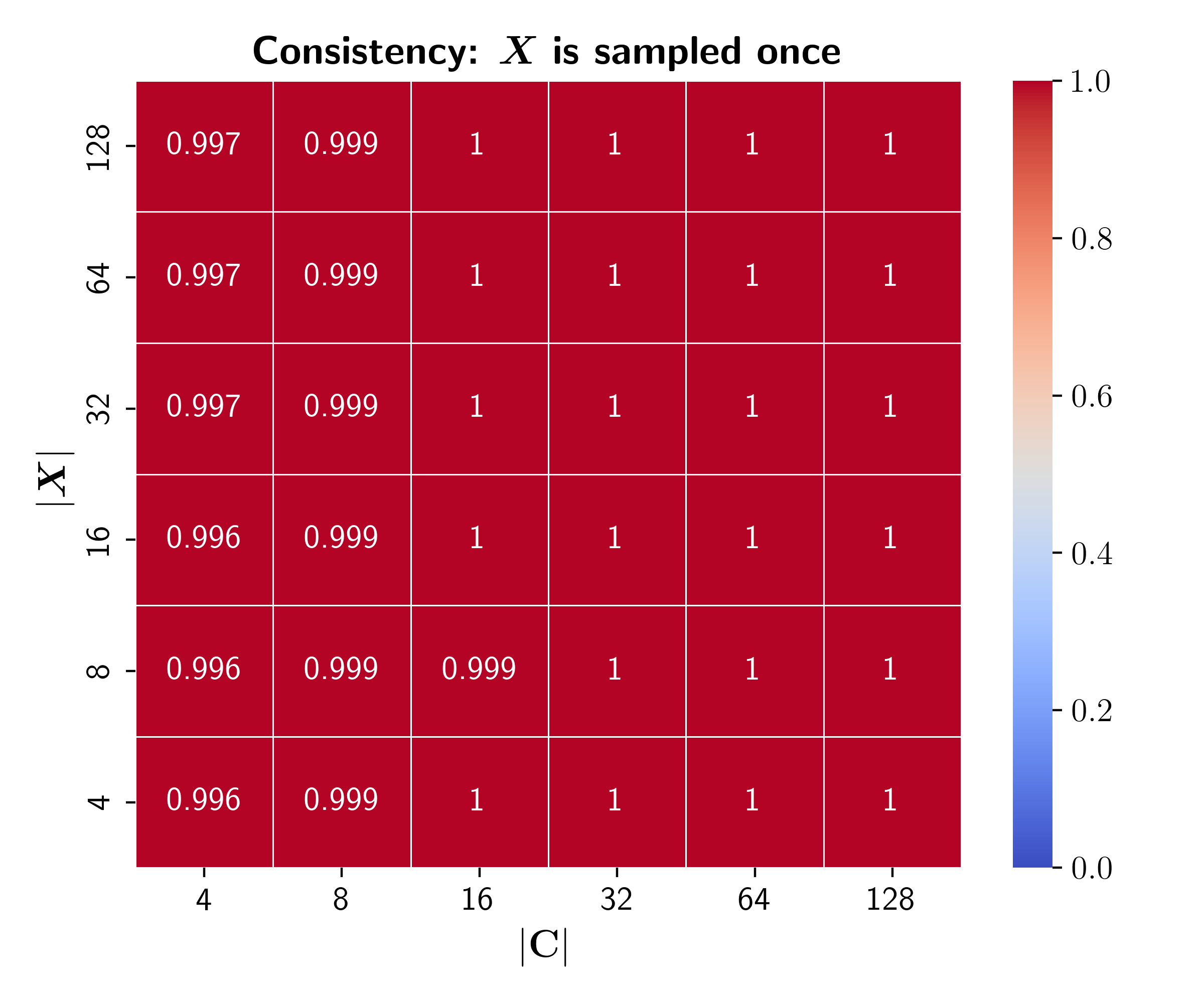} 
        \setcounter{subfigure}{2} 
        \caption{One $\ds$ sampled for all experiments.}
        \label{fig:consistency_all}
    \end{subfigure}
    \caption{A heatmap depicting the consistency of BED across different hyperparameter settings of the sampling parameters (the number of variable and constant values sampled, \#VS and \#CS, respectively) for the expressions with at most two variables.}
    \label{fig:consistency}
\end{figure}

Figure~\ref{fig:consistency} summarizes the results as a heatmap. As expected, consistency improves as $|\ds|$ and $|\mathbf{C}|$ increase. Moreover, we observe high correlations across rankings for most configurations, indicating substantial stability of the measure. Notably, random rankings would yield a correlation of approximately $0.02$, highlighting the robustness of the BED approximation. Based on these results, we set $|\ds|=64$ and $|\mathbf{C}|=32$ for subsequent experiments, as these values, according to the results reported in the graphs in Figure~\ref{fig:consistency} and \ref{app:consistency}, offer a favorable trade-off between consistency and computational cost.

\subsection{Clustering performance on equivalent expressions}\label{sec:clustering}
We next evaluate the effectiveness of BED in clustering behaviorally equivalent expressions. We hypothesize that equivalent expressions should have dissimilarity values close to zero, while non-equivalent expressions should be substantially more dissimilar. To test this, we generate sets of equivalent expressions, compute pairwise dissimilarities, cluster the resulting distance matrices, and evaluate the quality of the clusters. In addition, we visualize the distances using Multi-Dimensional Scaling (MDS)~\cite{cox2000multidimensional}.

To construct the evaluation dataset, we generate $16$ groups of behaviorally equivalent expressions, each containing $10$ expressions. For each group, we define a base expression and apply a diverse set of mathematical transformations to obtain equivalent variants. These transformations include algebraic properties (commutativity, associativity, and distributivity for addition and multiplication), simplification rules (e.g., $\frac{x}{x}=1$), and identities of elementary functions (e.g., $sin^2(x)+cos^2(x)=1$). The complete list of equivalent expression groups, specific transformations, and the description of the methodology used to generate equivalent expressions can be found in \ref{app:equivalent}.

To obtain robust results, we generate $10$ such datasets. For each dataset, we compute pairwise dissimilarities between all expressions using BED, edit distance, tree edit distance, and Jaro distance. Each resulting distance matrix is then used as input to hierarchical clustering~\cite{Sneath1973Hierarchical}, with the number of clusters fixed to $16$ (the known number of ground-truth groups). To address BED’s scale dependency, we additionally perform clustering using a column-normalized version of the BED distance matrix, treating its rows as feature vectors. This variant, denoted BED-CN, directly mitigates the issue that equivalence groups with high-magnitude outputs may exhibit larger within-group dissimilarities than low-magnitude groups, leading to distorted cluster structures.\footnote{Column-normalizing the distance matrices obtained using edit, tree edit, or Jaro distance has little effect on clustering outcomes; results are reported in the \ref{app:equivalent}.}

\begin{table}[ht]
\centering
\caption{Clustering performance of dissimilarity measures. Mean and standard deviation of Adjusted Rand Index (ARI), Silhouette score, V-measure, and Fowlkes-Mallows Index for various clustering methods. "BED as features" significantly outperforms other baselines.}\label{tab:clustering}
\resizebox{\linewidth}{!}{
\begin{tabular}{lcccc}
Baseline & ARI & Silhouette & V-measure & Fowlkes-Mallows \\
\midrule
Edit distance & $0.002$ ($\pm  0.001$) &  $0.318$ ($\pm  0.051$) &  $0.166$ ($\pm  0.011$) &  $0.217$ ($\pm  0.002$)\\
Tree edit distance & $0.004$ ($\pm  0.004$) &  $0.276$ ($\pm  0.060$) &  $0.184$ ($\pm  0.023$) &  $0.218$ ($\pm  0.003$)\\
Jaro &  $0.002$ ($\pm  0.002$) & $-0.120$ ($\pm  0.140$) &  $0.182$ ($\pm  0.019$) &  $0.213$ ($\pm  0.002$)\\
BED &   $0.203$ ($\pm  0.025$) &  $0.537$ ($\pm  0.014$) &  $0.697$ ($\pm  0.028$) &  $0.393$ ($\pm  0.018$)\\
BED-CN &  \bfseries $1.000$ ($\pm  0.000$) & \bfseries $0.939$ ($\pm  0.001$) & \bfseries $1.000$ ($\pm  0.000$) & \bfseries $1.000$ ($\pm  0.000$)\\ 
\end{tabular}}
\end{table}

We evaluate clustering performance using three external metrics---Adjusted Rand Index (ARI)~\cite{Hubert1985}, V-measure~\cite{rosenberg-hirschberg-2007-v}, and Fowlkes–Mallows Index~\cite{Fowlkes01091983}---which measure agreement with the ground truth, as well as the internal Silhouette score~\cite{ROUSSEEUW198753}, which assesses cluster compactness and separation. The results are shown in Table~\ref{tab:clustering}.

Syntactic measures (edit distance, tree edit distance, and Jaro distance) perform poorly, achieving external metric scores near zero and even a negative Silhouette score for Jaro distance, indicating a failure to group expressions by behavioral equivalence. BED, when used without normalization, substantially improves clustering quality (e.g., ARI of 0.203, V-measure of 0.697), but remains affected by scale dependency. In contrast, BED-CN achieves near-perfect performance across all metrics (ARI, V-measure, and Fowlkes–Mallows Index of 1) and a Silhouette score of $0.939$, demonstrating that BED captures behavioral similarity between expression far and significantly more effectively than syntax-based measures.

\begin{figure*}[ht]
\centering
\includegraphics[width=\linewidth]{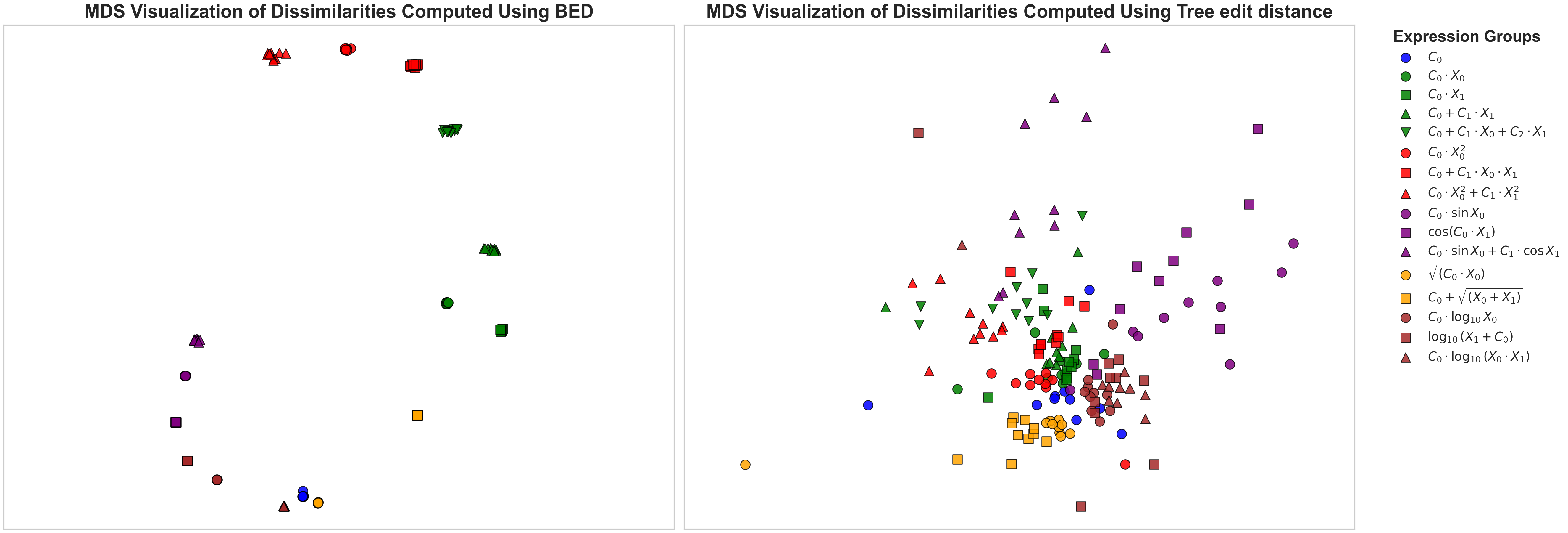}
\caption{Multi-Dimensional Scaling (MDS) visualizations of expression dissimilarities. The left plot uses distances computed with BED, while the right plot uses tree edit distance. Each color and shape combination represents a ground truth group of behaviorally equivalent expressions. Observe how BED effectively brings equivalent expressions into tight clusters, showcasing a logical arrangement of behavioral groups, in contrast to the scattered representation by tree edit distance.} \label{fig:clustering}
\end{figure*}

Figure~\ref{fig:clustering} visualizes the BED and tree edit distance matrices using MDS. The BED-based embedding, depicted in the graph on the left-hand side of the figure, forms tight, well-separated clusters of equivalent expressions (same color and shape), with a meaningful global organization of distinct behavioral groups in the embedding space. This indicates that BED captures both local equivalence and global behavioral structure. In sharp contrast, the tree edit distance embedding on the right-hand side of the figure is highly scattered, with substantial overlap between groups and no discernible structure, reflecting its focus on superficial syntactic form rather than behavior. Additional visualizations for edit and Jaro distances, which exhibit similar deficiencies, are provided in \ref{app:equivalent}.

\subsection{Smoothness of the error landscape}\label{sec:smoothness}
A key motivation for our proposed dissimilarity measure is its ability to induce a smooth error manifold over the space of mathematical expressions, particularly those with free parameters. For symbolic regression search, this property is essential: expressions that are close under the dissimilarity measure should have similar predictive error on a given dataset. In such a landscape, decreasing dissimilarity to the ground truth should reliably correspond to a decrease in error, enabling more efficient search. The central claim of this work is that BED substantially increases the smoothness of this error landscape.

To validate this claim, we experiment on $100$ datasets from the Feynman symbolic regression benchmark~\cite{Tegmark2020Feynman}. For each target equation from the benchmark, we randomly generate $100,000$ candidate expressions and compute their Root Mean Squared Error (RMSE) against the ground truth. We then calculate the dissimilarity between each candidate and the ground truth using BED, edit distance, tree edit distance, and Jaro distance, and rank the expressions according to each measure. Because the BED approximation is stochastic, we repeat its computation $10$ times and aggregate the results.

To examine the smoothness of the induced landscape, we focus on the top $n$ closest expressions for $n \in \{ 1, 2, \dots, 250 \}$. For each $n$, we compute the median RMSE of these expressions. The median is used instead of the mean due to its robustness to outliers, which are particularly frequent and problematic given the widely varying scale of the errors of the candidate expressions. Results using mean aggregation are included in \ref{app:error_manifold}.

\begin{figure*}[ht]
    \centering
    \begin{subfigure}[b]{0.48\textwidth}
        \centering
        \includegraphics[width=\textwidth]{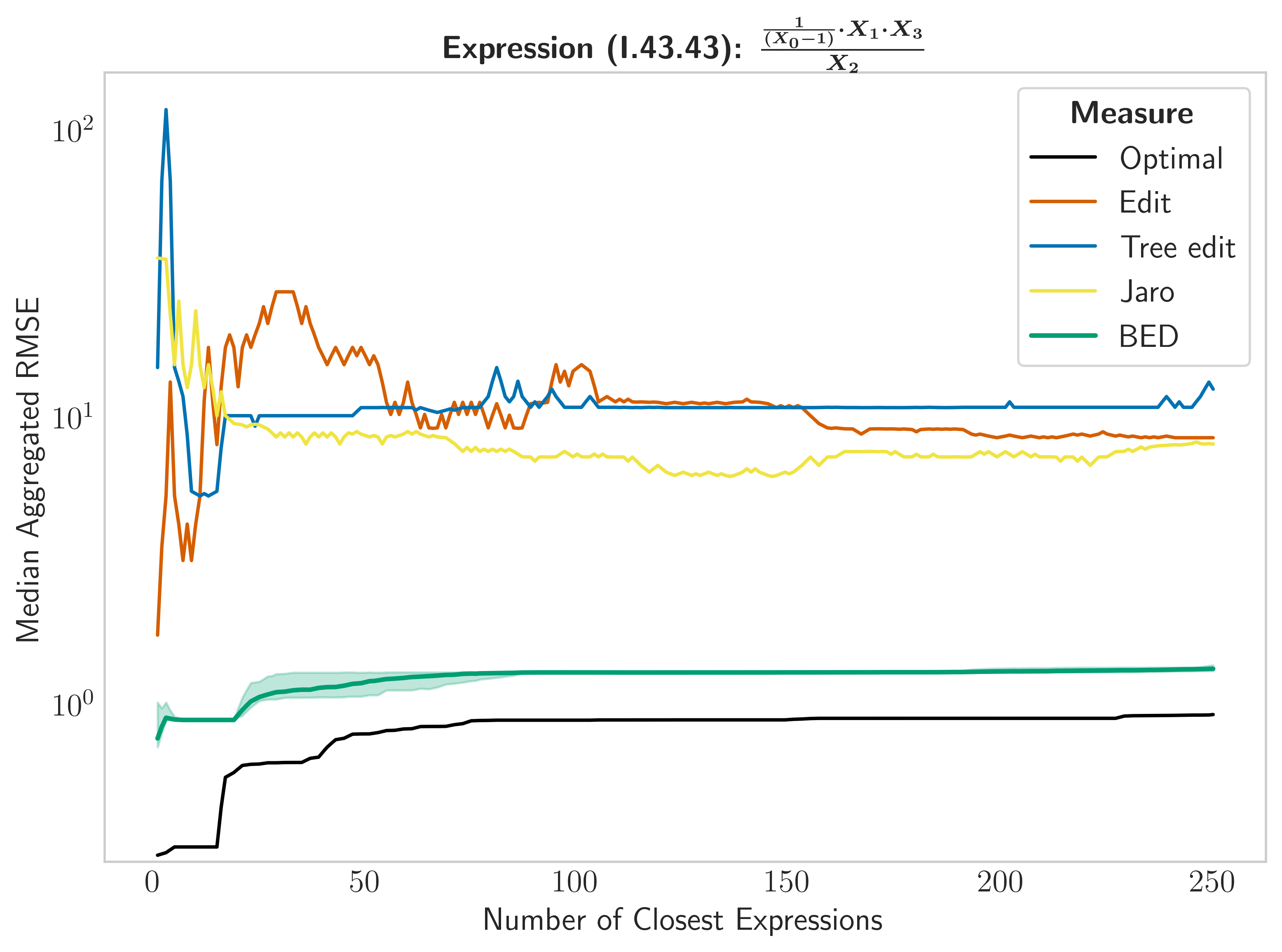} 
        \caption{Expression (I.43.43): $X_0 X_1 X_3 X_2^{-1}$}
        \label{fig:smoothness_a}
    \end{subfigure}
    \hfill
    \begin{subfigure}[b]{0.48\textwidth}
        \centering
        \includegraphics[width=\textwidth]{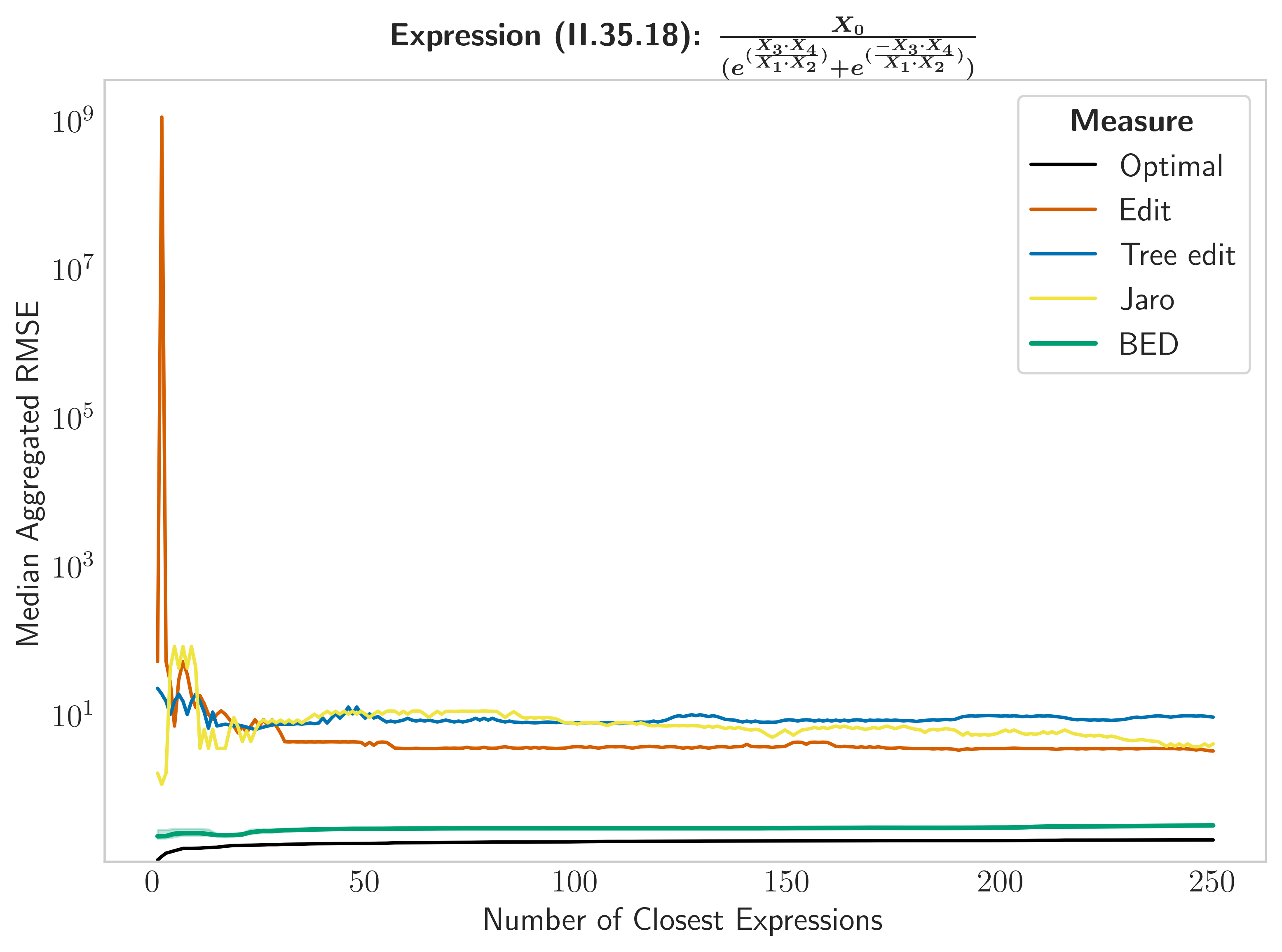} 
        \caption{Expression (II.35.18): $0.5 X_0\sech (\frac{X_3 X_4}{X_1 X_2})$}
        \label{fig:smoothness_b}
    \end{subfigure}
    \caption{Median aggregated RMSE for top closest expressions across two benchmark equations. Each subfigure illustrates how the median RMSE of the top N expressions, ordered by different dissimilarity measures, evolves as N increases (from 1 to 250). BED consistently achieves significantly lower and smoother aggregated RMSE curves compared to syntactic measures, demonstrating its superior ability to induce a smooth error manifold.}
    \label{fig:smoothness}
\end{figure*}

As a reference, we compare each measure to an Optimal oracle, which ranks expressions directly by their RMSE and selects the closest $250$. This oracle defines the lower bound of an ideally smooth error manifold.

Figure~\ref{fig:smoothness} shows that BED closely follows the Optimal curve, producing low and gradually increasing aggregated RMSE values with minimal variability (narrow confidence band). In contrast, the syntactic measures yield curves that are substantially higher and highly erratic, with frequent large fluctuations and early plateaus. This indicates that they fail to order expressions in a way that correlates with true error, and thus do not induce a smooth landscape for search.

To quantitatively evaluate performance across all $100$ Feynman datasets, we apply the following ranking procedure. For each dataset and each $n \in \{1, 2, \dots, 250\}$, we rank the four measures based on their median aggregated RMSE (where rank $1$ corresponds to the lowest median RMSE up to the point $n$ and rank $4$ to the highest). We then average these ranks across all $100$ datasets, yielding a single performance curve per measure.

\begin{figure}[ht]
\centering
\includegraphics[width=\linewidth]{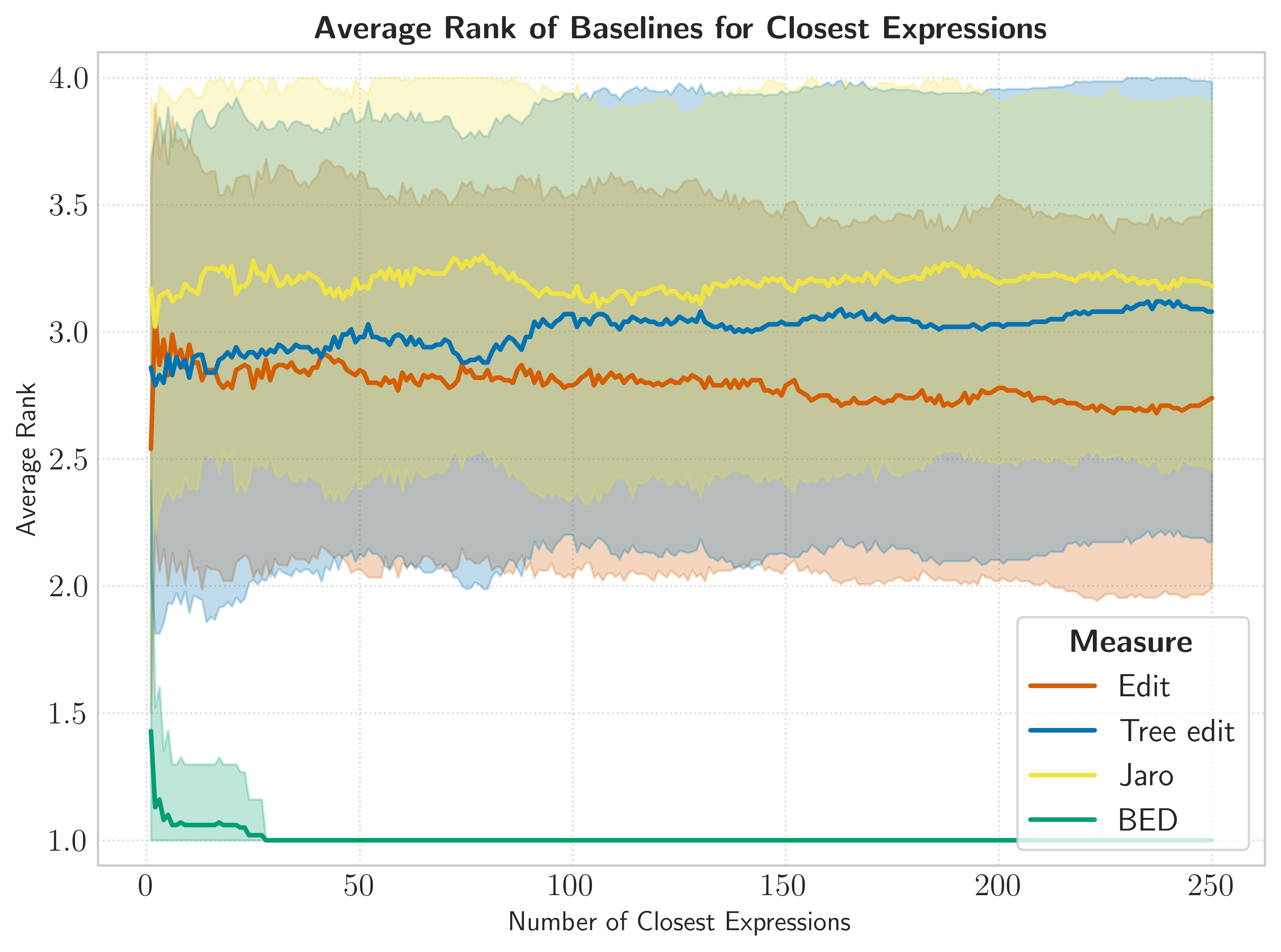}
\caption{Average Rank of Dissimilarity Measures for Ordering by RMSE. This figure shows the average rank of each dissimilarity measure (BED, edit distance, tree edit distance, Jaro distance) based on its ability to order expressions by their Root Mean Squared Error (RMSE) relative to a ground truth. At each point on the x-axis ("Number of Closest Expressions", n), the y-axis indicates the average rank of the measure whose top n expressions exhibit the lowest median RMSE. A lower average rank signifies better performance in creating a smooth error manifold.} \label{fig:ranking}
\end{figure}

The results, shown in Figure~\ref{fig:ranking}, demonstrate that BED maintains an average rank of approximately $1$ across all $n$, confirming that it consistently finds the lowest-error expressions. In contrast, edit distance, tree edit distance, and Jaro distance have average ranks between $2.5$ and $3.5$, indicating substantially worse performance. The shaded regions show the standard deviation, which is notably small for BED, highlighting the stability and consistency of its outstanding performance.

Finally, we note that the top $250$ closest expressions may include invalid expressions, which we skip in favor of the next closest valid ones. On average, edit distance yields $12.83$ invalid expressions within the top $250$, tree edit distance $7.61$, Jaro distance $11.61$, and BED only $4.66$. This further supports BED’s effectiveness at promoting meaningful, behaviorally valid expressions.

\section{Discussion}\label{sec:discussion}
\begin{figure*}[h!]
  \centering
  \includegraphics[width=\linewidth]{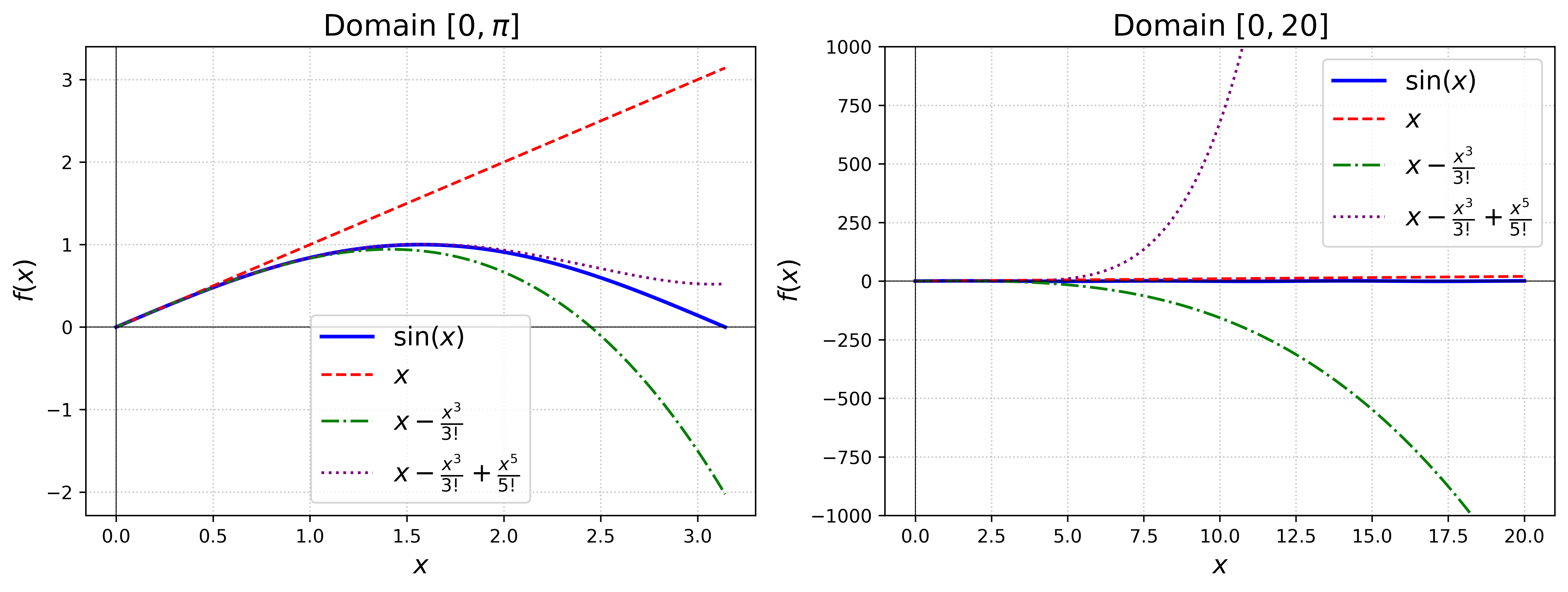}
  \caption{Comparison of $\sin x$ and its Taylor series approximations around $x = 0$ of increasing order over two different input domains of $[0, \pi]$ and $[0, 6 \pi]$.}
  \label{fig:sin}
\end{figure*}

One important aspect of behavior that we have not yet examined in detail is how the choice of the input domain affects behavioral similarity. Two expressions may exhibit nearly identical behavior within one domain, yet diverge significantly in another. This phenomenon is clearly demonstrated when comparing $\sin x$ to its Taylor series approximations around $x = 0$ of increasing order ($x$, $x-\frac{x^3}{6}$, $x-\frac{x^3}{6}+\frac{x^5}{120}$), as shown in Figure~\ref{fig:sin}. Syntactically, the edit distance between $\sin x$ and its polynomial approximations grows monotonically with the order of expansion and is independent of the domain.

\begin{table}[htbp]
    \centering
    \caption{Dissimilarity between $\sin x$ and its Taylor expansions around $x=0$ of increasing order, computed over different input domains.}
    \label{tab:sin_dissimilarity}
    \begin{tabular}{lrrr}
        Expansion & Edit Distance & RMSE ($[0, \pi]$) & RMSE ($[0, 6 \pi]$) \\
        \midrule
        $x$ & 1 & 1.34 & 11.60 \\
        $x-\frac{x^3}{6}$ & 7 & 0.63 & 493.99 \\[0.4em] 
        $x-\frac{x^3}{6}+\frac{x^5}{120}$ & 14 & 0.14 & 7576.69 \\[0.4em] 
    \end{tabular}
\end{table}

Behaviorally, however, the degree of similarity depends strongly on the interval over which the functions are compared. Within the domain $[0, \pi]$, higher-order expansions provide a closer match to $\sin x$, reflecting the local accuracy of the Taylor expansion near the expansion point. In contrast, over a wider domain such as $[0, 6 \pi]$, the higher-order polynomial terms grow rapidly in magnitude and produce large oscillations that deviate from the bounded, periodic behavior of $\sin x$, which remains confined to the interval $[-1, 1]$. Consequently, lower-order expansions may, on average, exhibit smaller deviations from $\sin x$ across the extended domain, even though they are less accurate near $x = 0$. This dependence of behavioral dissimilarity on the input domain is quantitatively illustrated in Table~\ref{tab:sin_dissimilarity}, which reports RMSE values computed over $1000$ equidistant points in each domain.

Another critical aspect of behavioral analysis concerns the applicability of behavior-based dissimilarity measures to mathematical expressions containing free parameters. A key advantage of syntactic measures, such as edit distance, lies in their simplicity and computational efficiency, which makes them well-suited for generating new, syntactically similar expressions. This efficiency, however, represents a major challenge for BED, currently the only behavior-based dissimilarity measure applicable to expressions with free parameters. As such, BED poses certain limitations for traditional, non-neural symbolic regression methods.

Nevertheless, BED holds substantial promise for neural-based symbolic regression. Its formulation enables models to capture intricate behavioral relationships between parameterized expressions more effectively. While one can compare such expressions by fitting their parameters and evaluating them on a shared dataset using metrics like RMSE, this approach implicitly assumes that the model can always identify and exploit the optimal parameters when determining relationships among expressions. In other words, the model’s latent space must indirectly encode these fitted parameters to accurately represent behavioral similarity. BED, in contrast, does not rely on such assumptions. It naturally treats parameters as distributions, allowing behavioral relationships to be established without requiring explicit parameter fitting. This property should yield a more consistent and robust measure of similarity between expressions, making BED a valuable tool for neural architectures in symbolic regression.

Beyond model training, BED also offers promising applications in benchmarking symbolic regression algorithms. Existing benchmarks typically focus on reconstructing deterministic target expressions—that is, expressions with specific constant values. Such evaluation often overemphasizes low RMSE values, favoring overparameterized expressions that overfit the data. By contrast, a benchmark based on BED could define the ground truth using parameterized expressions and measure performance based on how similarly these parameterized expressions behave. This would encourage models to recover the underlying functional form rather than merely fitting numerical parameters, thereby reducing overfitting and providing a more faithful assessment of symbolic regression performance.

Despite its advantages, BED has several limitations. The first stems from the intractability of Eq.~\eqref{eq:bed}, which necessitates a stochastic approximation. Although our experiments show that this approximation is highly consistent, its probabilistic nature may introduce minor variability in the computed dissimilarities. Moreover, BED is computationally demanding: sampling over parameter and input spaces, evaluating expressions, and computing Wasserstein distances collectively make it substantially more expensive than simple syntax-based measures such as edit distance, with a computational cost comparable to that of more complex measures like tree edit distance.

A second limitation arises from the scale dependence of BED. Expressions producing outputs of larger magnitude tend to yield higher BED values. This behavior is evident in Section~\ref{sec:clustering}, where expressions equivalent to $C\cdot X_0^2 + C\cdot X_1^2$ are behaviorally closer to each other than to any other expressions, yet remain farther apart than groups of expressions equivalent to $C$ and $\sqrt{C\cdot X_0}$. Consequently, the clustering algorithm first merges groups corresponding to $C$ and $\sqrt{C\cdot X_0}$, and only later those corresponding to $C\cdot X_0^2 + C\cdot X_1^2$.

Finally, BED behaves differently when comparing an expression with free parameters to its fitted counterpart. Although smaller BED values indicate greater behavioral similarity to the ground truth, the dissimilarity between the true expression and its version with free parameters is never exactly zero. This effect is visible in Figure~\ref{fig:smoothness}, where BED approaches but rarely coincides with the optimal oracle—particularly for the closest matches, where the oracle value approaches zero. Consequently, BED should not be interpreted as a direct convergence criterion toward zero dissimilarity; rather, it should serve as a guide toward the minimum attainable dissimilarity, corresponding to the highest behavioral similarity achievable within the space of expressions with free parameters.

While we focused on quantifying behavioral dissimilarity between mathematical expressions with free parameters, the underlying framework of BED is far more broadly applicable to a variety of algorithmic and computational models. The core principle—recasting a parameterized function as a joint probability distribution over its inputs and outputs— extends naturally beyond symbolic expressions. For instance, machine learning models can be viewed as parametrized expressions, where hyperparameters (such as the regularization constant $C$ in an SVM, or the number of trees in a random forest) play the role of free parameters, and both training and test data define the input domain $\mathcal{X}$. Applying the BED formulation in this context would yield a parameter-aware distance measure between models, enabling the clustering of models with similar behavior, the comparison of optimization strategies, and the structuring of large model collections based on functional similarity, independent of model architecture.

\section{Conclusion}\label{sec:conclusion}
This paper introduces Behavior-aware Expression Dissimilarity (BED), a behavior-based measure for quantifying the dissimilarity between mathematical expressions containing free parameters. Such expressions describe families of functions and are central to many computational settings, including but not limited to symbolic regression, where the goal is to efficiently explore the space of candidate expressions while optimizing constants against data after selection. In Section~\ref{sec:behavior}, we represent expressions as joint probability distributions over their input and output values, inherently capturing the variability in output introduced by free parameters. This conceptual shift enables the use of probabilistic tools---specifically, the Wasserstein distance---to quantify behavioral dissimilarity, as formalized in Section~\ref{sec:definition}. Because the measure is analytically intractable, Section~\ref{sec:approximation} presents a stochastic approximation scheme, and Section~\ref{sec:consistency} demonstrates that this approximation is highly robust, yielding consistent and stable results across a range of hyperparameter configurations.

The central hypothesis of this work is that BED provides a more meaningful and interpretable structuring of the space of mathematical expressions than traditional syntax-based measures. The clustering experiments in Section~\ref{sec:clustering} strongly support this hypothesis. Whereas syntactic measures (edit, tree edit, and Jaro distance) yielded clustering scores close to random, BED achieved near-perfect performance across all external metrics, successfully grouping behaviorally equivalent expressions. Furthermore, the smoothness analysis in Section~\ref{sec:smoothness} confirms that proximity of expressions in the BED-induced space reliably corresponds to reductions in their prediction error. Across benchmarks, BED consistently maintains an average rank of $1$, demonstrating its ability to induce a smoother, behaviorally coherent structure over the space of expressions---substantially outperforming all syntax-based alternatives.

Beyond its empirical validation, BED provides a general foundation for behavior-based comparison, clustering, and learning of equations from data. It offers a principled way to relate expressions that are syntactically different but behaviorally similar, enabling more efficient and interpretable exploration of functional relationships. Ongoing work integrates BED into symbolic regression and generative modeling frameworks, where a BED-based regularization term structures latent spaces so that nearby points correspond to behaviorally similar expressions. Such structuring facilitates smoother optimization and more efficient exploration of the expression space, supporting optimization algorithms that exploit behavioral smoothness.

Another promising direction is the development of a new benchmark for symbolic regression grounded in BED. As discussed, existing benchmarks tend to reward overfitting by focusing on the reconstruction of deterministic, fully fitted expressions. In contrast, a BED-based benchmark would employ expressions with free parameters as the ground truth and evaluate models using behavioral dissimilarity. This paradigm shifts the emphasis from minimizing RMSE between fitted outputs to identifying the correct underlying functional structure, thereby mitigating overfitting and enabling a more meaningful and robust evaluation of symbolic regression methods.

\section*{CRediT authorship contribution statement}
\textbf{Sebastian Mežnar}: Conceptualization, Data curation, Formal analysis, Investigation, Methodology, Visualization, Software, Writing – original draft, Writing – review and editing; \textbf{Sašo Džeroski}: Funding acquisition, Supervision, Validation, Writing – review and editing; \textbf{Ljupčo Todorovski}: Conceptualization, Funding acquisition, Methodology, Supervision, Validation, Writing – original draft, Writing – review and editing.

\section*{Declaration of competing interest}
The authors declare that they have no known competing financial interests or personal relationships that could have appeared to influence the work reported in this paper.

\section*{Acknowledgments}
This research was funded by the Slovenian Research Agency via the research program ``Knowledge Technologies" (P2-0103), the Gravity project ``AI for science" (GC-0001), and by the ARRS Grant for young researchers (first author). The authors especially appreciate the helpful comments and suggestions by Nikola Simidjievski and the fruitful discussions within the SHED discussion group (with Jure Brence, Boštjan Gec, Nina Omejc, and Martin Perčinić).

\section*{Data availability}
All data used for the experiments is either synthetically generated following the procedures detailed within the paper or publicly available and cited. The complete codebase for the Behavior-based Dissimilarity (BED) measure and the synthetic data generation scripts is publicly available at a dedicated repository to ensure full reproducibility.

\section*{Declaration of generative AI and AI-assisted technologies in the manuscript preparation process}
During the preparation of this work the authors used Google's Gemini and OpenAI's ChatGPT in order to refine the text in the manuscript. After using this tool/service, the authors reviewed and edited the content as needed and take full responsibility for the content of the published article.

\appendix
\section{Mathematical properties of the proposed measures}\label{app:characteristics}
This appendix details the mathematical properties of the continuous integral form of Behavior-aware Expression Dissimilarity (BED) measure from Section~\ref{sec:definition}. Let's start with the proof of non-negativity.

\begin{proposition}[Non-Negativity]
The BED measure is non-negative, i.e., $\text{BED}^{(\mathcal{X})}(\mathit{E},\mathit{F}) \ge 0$.
\end{proposition}
\begin{proof}
We will show the non-negativity of BED by first proving that the value of BED for a single input point is always non-negative, and then extending this property to the whole domain.

The non-negativity of the single-point BED, $\text{BED}^{(\dv)}(\mathit{E},\mathit{F})$, can be proven by considering all possible scenarios at a given input point $\dv$:
\begin{enumerate}
    \item When both expressions are undefined at $\dv$ their dissimilarity is set to $0$ by definition, which is a non-negative value.
    \item When only one expression is defined at $\dv$ their dissimilarity is set to $+\infty$ by definition, which is a non-negative value.
    \item When both expressions are defined at $\dv$, $\text{BED}^{(\dv)}(\mathit{E},\mathit{F})$ is defined by the integral of the absolute difference of their quantile functions. The integrand is an absolute value and thus always non-negative. As the integral of a non-negative function over a non-empty domain is also non-negative, $\text{BED}^{(\dv)}(\mathit{E},\mathit{F}) \ge 0$.
\end{enumerate}
In all possible scenarios, the single-point BED measure is non-negative.

Next, let's examine BED across the entire input domain. Here, BED is defined as the integral of the single-point measure, normalized by the volume of the domain:
$$\text{BED}^{(\mathcal{X})}(\mathit{E},\mathit{F}) = \frac{1}{\text{Vol}(\mathcal{X})} \int_{\mathcal{X}} \text{BED}^{(\dv)}(\mathit{E},\mathit{F}) \, d\mathcal{X}$$

From the first part of the proof, we know that the integrand, $\text{BED}^{(\dv)}(\mathit{E},\mathit{F})$, is non-negative for all $\dv \in \mathcal{X}$. Since the normalizing constant $\frac{1}{\text{Vol}(\mathcal{X})}$ is also positive, the result of the integral and subsequent multiplication must be non-negative.
\end{proof}

Next, we show that the dissimilarity of an expression to itself is zero.

\begin{proposition}[Reflexivity]
The dissimilarity of an expression to itself is zero, i.e., $\text{BED}^{(\dd)}(\mathit{E},\mathit{E}) = 0$.
\end{proposition}
\begin{proof}
We prove this property by first establishing that the single-point BED is zero and then extending this result to the whole domain.

We establish that the single-point BED of an expression to itself is zero, $\text{BED}^{(\dv)}(\mathit{E},\mathit{E})$, by considering all possible scenarios at a given input point $\dv$:
\begin{enumerate}
    \item When expression $E$ is defined at $\dv$: The single-point BED of the expression to itself is defined as:
    $$\text{BED}^{(\dv)}(\mathit{E},\mathit{E}) = \int_0^1 \left |Q_{P_{\mathit{E}}(\dv)}(q)-Q_{P_{\mathit{E}}(\dv)}(q)\right | \, dq .$$
    The integrand is the absolute value of the difference between identical quantile functions, which is zero for all $q$. The integral of zero is zero.
    \item When expression $E$ is undefined at $\dv$: By definition, the dissimilarity of an expression to itself (or, in other words, between two expressions undefined at $\dv$) is set to 0.
\end{enumerate}
In both cases, the single-point BED measure of an expression to itself is zero.

This property extends to the overall BED measure, which is defined as the integral of the single-point measure normalized by the volume of the domain:
$$\text{BED}^{(\mathcal{X})}(\mathit{E},\mathit{E}) = \frac{1}{\text{Vol}(\mathcal{X})} \int_{\mathcal{X}} \text{BED}^{(\dv)}(\mathit{E},\mathit{E}) \, d\mathcal{X}$$
Since the integrand, $\text{BED}^{(\dv)}(\mathit{E},\mathit{E})$, is zero for all $\dv \in \mathcal{X}$, the integral of zero is zero, and a zero multiplied by a non-zero constant remains zero. Therefore, $\text{BED}^{(\mathcal{X})}(\mathit{E},\mathit{E}) = 0$.
\end{proof}

Now we prove that the BED measure is symmetric.

\begin{proposition}[Symmetry]
The BED measure is symmetric, i.e., $\text{BED}^{(\dd)}(\mathit{E},\mathit{F}) = \text{BED}^{(\dd)}(\mathit{F},\mathit{E})$.
\end{proposition}
\begin{proof}
We first show that a single-point BED is symmetric and then extend this to the whole domain.

For the single-point BED, $\text{BED}^{(\dv)}(\mathit{E},\mathit{F})$, we have the following cases:
\begin{enumerate}
    \item When both expressions are undefined at $\dv$, $\text{BED}^{(\dv)}(\mathit{E},\mathit{F}) = 0$ by definition. Since the order of $\mathit{E}$ and $\mathit{F}$ does not affect the definition, $\text{BED}^{(\dv)}(\mathit{F},\mathit{E}) = 0 = \text{BED}^{(\dv)}(\mathit{E},\mathit{F})$.
    \item When only one expression is defined at $\dv$, the dissimilarity is set to $\infty$. The order of the expressions does not affect this definition, so $\text{BED}^{(\dv)}(\mathit{E},\mathit{F}) = +\infty = \text{BED}^{(\dv)}(\mathit{F},\mathit{E})$.
    \item When both expressions are defined at $\dv$, the single-point BED is defined by the integral of the absolute difference of their quantile functions:
    $$\text{BED}^{(\dv)}(\mathit{E},\mathit{F}) = \int_0^1 \left |Q_{P_{\mathit{E}}(\dv)}(q)-Q_{P_{\mathit{F}}(\dv)}(q)\right | \, dq$$
    Using the property of absolute values, $\left |a-b\right | = \left | b-a\right |$, we can rewrite the integrand:
    $$\left |Q_{P_{\mathit{E}}(\dv)}(q)-Q_{P_{\mathit{F}}(\dv)}(q)\right | = \left |Q_{P_{\mathit{F}}(\dv)}(q)-Q_{P_{\mathit{E}}(\dv)}(q)\right |$$
    Substituting this back into the integral, we get the definition of $\text{BED}^{(\dv)}(\mathit{F},\mathit{E})$:
    $$\text{BED}^{(\dv)}(\mathit{E},\mathit{F}) = \int_0^1 \left |Q_{P_{\mathit{F}}(\dv)}(q)-Q_{P_{\mathit{E}}(\dv)}(q)\right | \, dq = \text{BED}^{(\dv)}(\mathit{F},\mathit{E})$$
\end{enumerate}
This shows that the single-point BED measure is symmetric is all cases.

Next, let's look at the overall definition of BED. BED is defined as:
$$\text{BED}^{(\mathcal{X})}(\mathit{E},\mathit{F}) = \frac{1}{\text{Vol}(\mathcal{X})} \int_{\mathcal{X}} \text{BED}^{(\dv)}(\mathit{E},\mathit{F}) \, d\mathcal{X}$$From the first part, we know that the integrand is symmetric, i.e., $\text{BED}^{(\dv)}(\mathit{E},\mathit{F}) = \text{BED}^{(\dv)}(\mathit{F},\mathit{E})$. We can substitute this into the overall formula:$$\text{BED}^{(\mathcal{X})}(\mathit{E},\mathit{F}) = \frac{1}{\text{Vol}(\mathcal{X})} \int_{\mathcal{X}} \text{BED}^{(\dv)}(\mathit{F},\mathit{E}) \, d\mathcal{X} = \text{BED}^{(\mathcal{X})}(\mathit{F},\mathit{E})$$
Thus, the overall BED measure is also symmetric.
\end{proof}

Next, we examine the triangle inequality.

\begin{proposition}[Triangle Inequality]
The BED measure satisfies the triangle inequality $\text{BED}^{(\dd)}(\mathit{E},\mathit{F}) \le \text{BED}^{(\dd)}(\mathit{E},\mathit{G}) + \text{BED}^{(\dd)}(\mathit{G},\mathit{F})$.
\end{proposition}
\begin{proof}
We first show that the triangle inequality holds for the single-point BED and then extend this to the whole domain.

Let $\mathcal{X}$ be a non-empty input domain and $\dv_0 \in \mathcal{X}$ be a single point. Let $D_{\mathit{E}}$, $D_{\mathit{F}}$, and $D_{\mathit{G}}$ be binary indicators (where $1$ means "defined" and $0$ means "undefined") for whether expressions $\mathit{E}$, $\mathit{F}$, and $\mathit{G}$ are defined at point $\dv_0$.

We examine all possible scenarios in Table~\ref{tab:triangle_inequality}. When both expressions are defined at $\dv_0$, the distance is given by the $1$-Wasserstein distance, which we denote as $W_1(\cdot, \cdot)$. The triangle inequality, $\text{BED}^{(\dv_0)}(\mathit{E}, \mathit{G}) \le \text{BED}^{(\dv_0)}(\mathit{E}, \mathit{F}) + \text{BED}^{(\dv_0)}(\mathit{F}, \mathit{G})$, holds trivially for all rows in the table except the final one (row 8), as any scenario involving $\infty$ on the right hand side ensures the inequality is satisfied. The inequality also holds for the final row, as the single-point BED simplifies to the $1$-Wasserstein distance, for which the triangle inequality property holds.

\begin{table}[htbp]
\centering
\caption{Validation of the triangle inequality for the single-point $\text{BED}^{(\dv_0)}$. The table examines all eight combinations of definition status ($D_{\cdot}=1$ for defined, $D_{\cdot}=0$ for undefined) for three expressions $\mathit{E}, \mathit{F}, \mathit{G}$ and the distances between these expressions at point $\dv_0$.}
\label{tab:triangle_inequality}
\begin{tabular}{cccccc}
$D_{\mathit{E}}$ & $D_{\mathit{F}}$ & $D_{\mathit{G}}$ & $\text{BED}^{(\dv)}(\mathit{E},\mathit{F})$ & $\text{BED}^{(\dv)}(\mathit{E},\mathit{G})$ & $\text{BED}^{(\dv)}(\mathit{G},\mathit{F})$ \\
\midrule
$0$ & $0$ & $0$ & $0$                   & $0$                   & $0$                   \\
$0$ & $0$ & $1$ & $0$                   & $\infty$              & $\infty$              \\
$0$ & $1$ & $0$ & $\infty$              & $0$                   & $\infty$              \\
$0$ & $1$ & $1$ & $\infty$              & $\infty$              & $W_1(\mathit{G},\mathit{F})$   \\
$1$ & $0$ & $0$ & $\infty$              & $\infty$              & $0$                   \\
$1$ & $0$ & $1$ & $\infty$              & $W_1(\mathit{E},\mathit{G})$   & $\infty$              \\
$1$ & $1$ & $0$ & $W_1(\mathit{E},\mathit{F})$   & $\infty$              & $\infty$              \\
$1$ & $1$ & $1$ & $W_1(\mathit{E},\mathit{F})$   & $W_1(\mathit{E},\mathit{G})$   & $W_1(\mathit{G},\mathit{F})$   \\
\end{tabular}
\end{table}

Therefore, for any input point $\dv$:
$$\text{BED}^{(\dv)}(\mathit{E},\mathit{G}) \le \text{BED}^{(\dv)}(\mathit{E},\mathit{F}) + \text{BED}^{(\dv)}(\mathit{F},\mathit{G}) .$$

Next, the overall BED measure is defined as:
$$\text{BED}^{(\mathcal{X})}(\mathit{E},\mathit{G}) = \frac{1}{\text{Vol}(\mathcal{X})} \int_{\mathcal{X}} \text{BED}^{(\dv)}(\mathit{E},\mathit{G}) \, d\mathcal{X} .$$From the first part, we know that the inequality $\text{BED}^{(\dv)}(\mathit{E},\mathit{G}) \le \text{BED}^{(\dv)}(\mathit{E},\mathit{F}) + \text{BED}^{(\dv)}(\mathit{F},\mathit{G})$ holds for all $\dv \in \mathcal{X}$. We can integrate this inequality over the domain $\mathcal{X}$ and preserve the order:$$\int_{\mathcal{X}} \text{BED}^{(\dv)}(\mathit{E},\mathit{G}) \, d\mathcal{X} \le \int_{\mathcal{X}} \left( \text{BED}^{(\dv)}(\mathit{E},\mathit{F}) + \text{BED}^{(\dv)}(\mathit{F},\mathit{G}) \right) \, d\mathcal{X} .$$Using the linearity of integrals and normalizing by the positive constant $\frac{1}{\text{Vol}(\mathcal{X})}$, we arrive at the triangle inequality for the overall measure:$$\text{BED}^{(\mathcal{X})}(\mathit{E},\mathit{G}) \le \text{BED}^{(\mathcal{X})}(\mathit{E},\mathit{F}) + \text{BED}^{(\mathcal{X})}(\mathit{F},\mathit{G}) .$$
\end{proof}

Lastly, the identity of indiscernibles axiom—that the dissimilarity between two items is zero if and only if they are identical—requires careful consideration when applied to the BED measure. Its validity depends entirely on whether we view expressions as symbolic objects or as probability distributions.

On the one hand, if we consider expressions as symbolic entities (e.g., as expression trees or sequences of symbols), the axiom does not hold. For example, expressions such as $X_0 + X_1$ and $X_1 + X_0$ are syntactically distinct objects, yet they are behaviorally equivalent. Consequently, the BED dissimilarity between them is zero, thus violating the identity of indiscernibles.

On the other hand, the BED measure is fundamentally defined on the behavior of expressions, as captured by their output probability distributions. When viewed at this level of abstraction, the axiom holds true. Since the dissimilarity is zero if and only if the underlying distributions are identical, we can conclude that the identity of indiscernibles axiom is valid for the distributions themselves.

\section{Probabilistic context-free grammar for generating random expressions}\label{app:grammar}
In each of the experiments evaluating the BED approximation, we generate random expressions by sampling the following probabilistic context-free grammar~\cite{Brence2021ProGED}: 
\begin{align*}
    E &\to E + F \, [0.2004] \; \vert \; E - F \, [0.1108] \; \vert \; F \, [0.6888] \\
    F &\to F \cdot T \, [0.3349] \; \vert \; \frac{F}{T} \, [0.1098] \; \vert \; T \, [0.5553] \\
    T &\to C \, [0.1174] \; \vert \; R \, [0.1746] \; \vert \; X \, [0.708] \\
    R &\to (E) \, [0.6841] \; \vert \; E^P \, [0.0036] \; \vert \; \sin (E) \, [0.028] \; \vert \\
      & \cos (E)\, [0.049] \; \vert \; \sqrt{E} \, [0.0936] \; \vert \; \exp (E) \, [0.0878] \; \vert \\
      & \log (E) \, [0.0539] \\
    P &\to 2 \, [0.65] \; \vert \; 3 \, [0.35] .
\end{align*}
In this grammar, $C$ represents a free parameter, and $X$ is a placeholder symbol for a variable. When an expression is generated, this $X$ symbol is uniformly sampled to represent a variable from a predefined set of possible variables. The grammar itself is derived from the standard context-free grammar of mathematical expressions, containing frequently used operators and functions. The probabilities for its production rules were determined by maximizing the likelihood of mathematical expressions included in a corpus extracted from Wikipedia~\cite{Guimera2020BayesianScientist,Marija2022Probs}.

\section{Additional results for measuring the impact of sampling on BED approximation}\label{app:consistency}
In Section~\ref{sec:consistency}, we established the consistency of BED approximations for expressions involving up to two variables. Here, we present supplementary experiments that further confirm the robustness of the measure under broader conditions.

\begin{figure}[h!]
     \centering
     \begin{subfigure}[b]{0.49\linewidth}
         \centering
         \includegraphics[width=\textwidth]{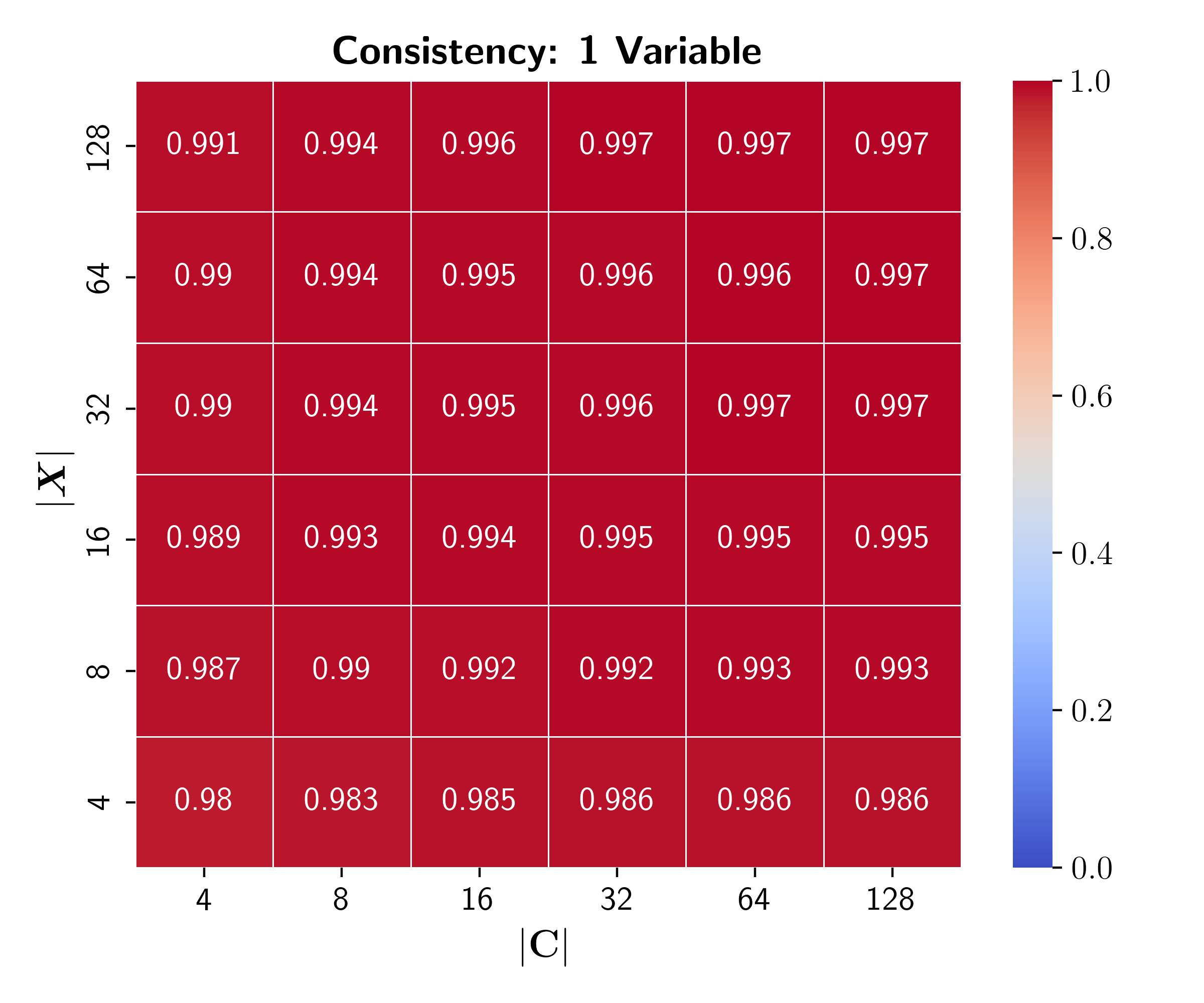}
     \end{subfigure}
     \hfill
     \begin{subfigure}[b]{0.49\linewidth}
         \centering
         \includegraphics[width=\textwidth]{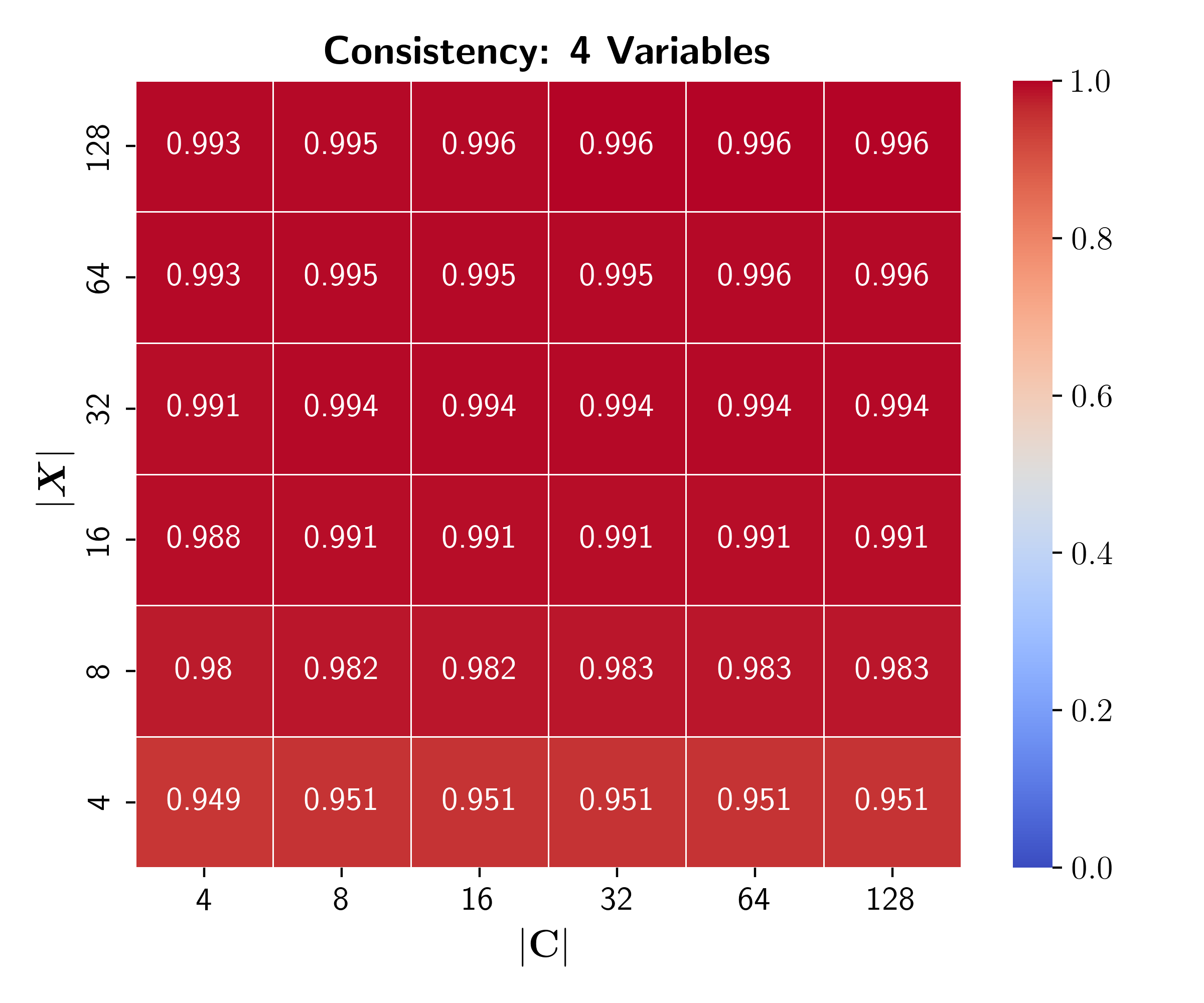}
     \end{subfigure}
     
     \begin{subfigure}[b]{0.49\linewidth}
         \centering
         \includegraphics[width=\textwidth]{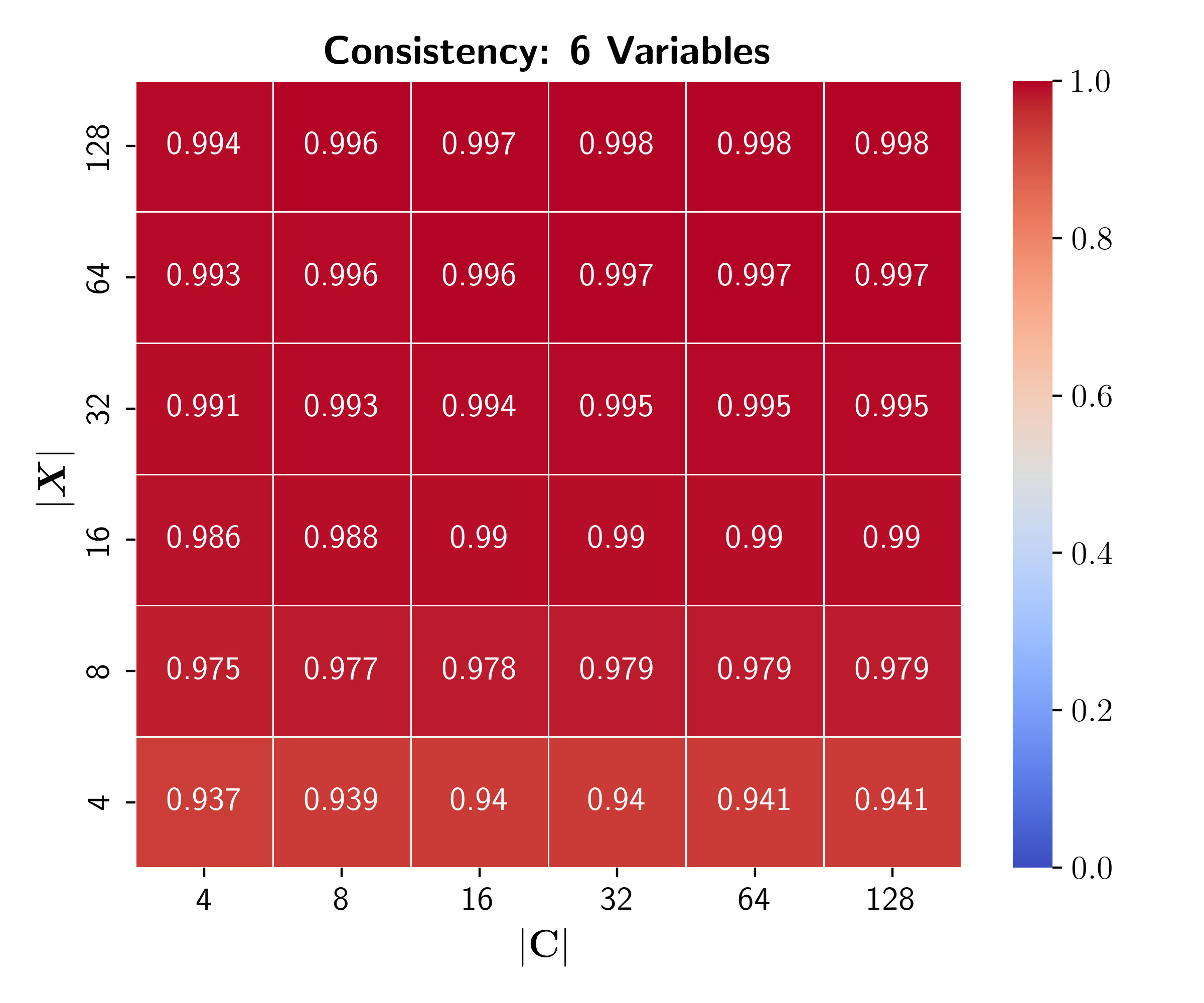}
     \end{subfigure}
     \hfill
     \begin{subfigure}[b]{0.49 
     \linewidth}
         \centering
         \includegraphics[width=\textwidth]{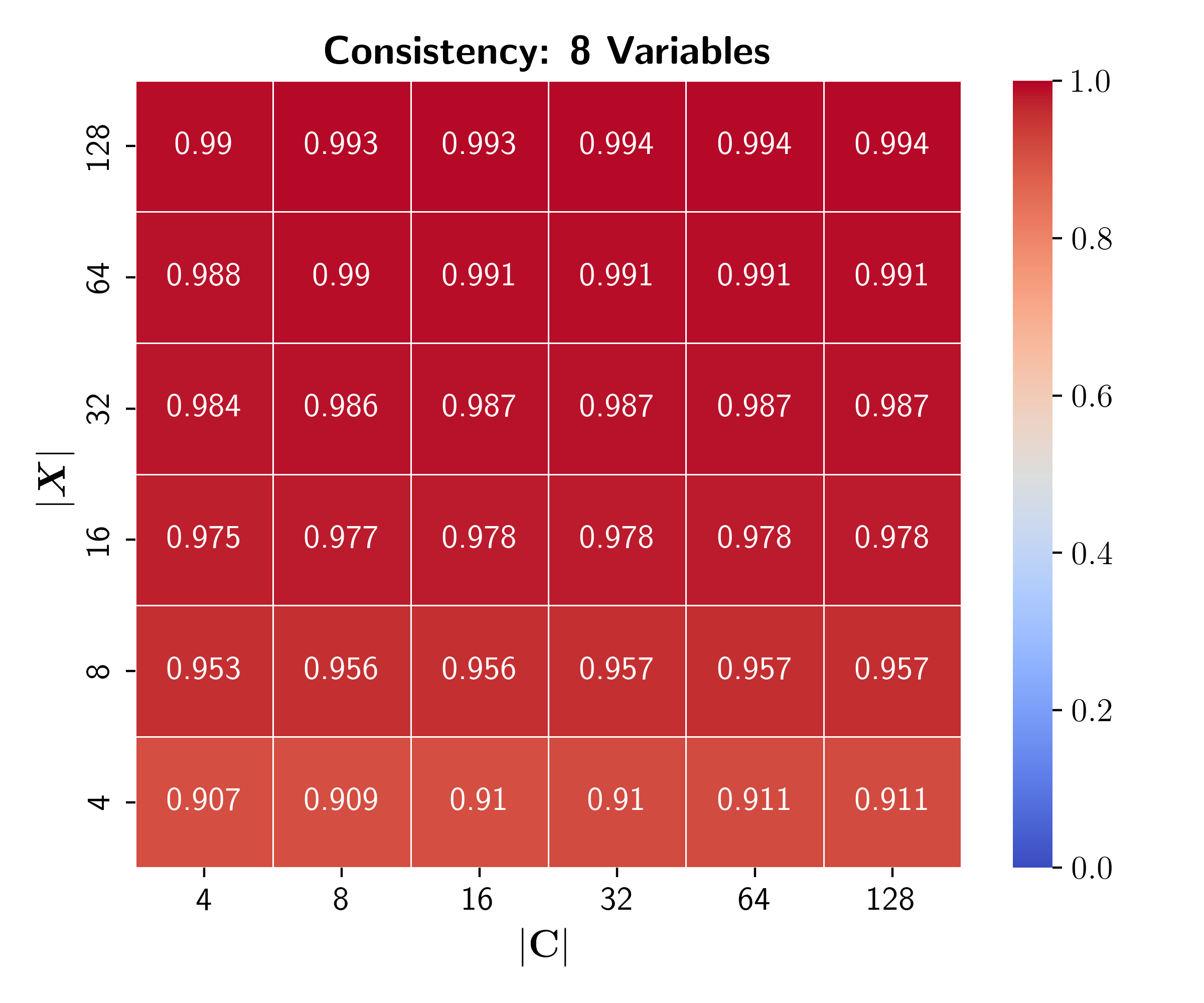}
     \end{subfigure}
\caption{Heatmaps showing the consistency of BED across different settings of the sampling parameters (number of variable and constant values sampled, $|\ds|$ and $|\mathbf{C}|$) for the expressions with varying number of variables (check the labels on top of each heatmap).}\label{fig:app_consistency}
\end{figure}

We investigate the influence of the number of variables using the same experimental setup as in Section~\ref{sec:consistency}. Figure~\ref{fig:app_consistency} reports results for sets of expressions containing up to $1$, $4$, $6$, and $8$ variables. As observed in Figure~\ref{fig:consistency}, the BED approximation exhibits strong correlation across rankings for most hyperparameter configurations. Importantly, this level of consistency remains effectively unchanged as the dimensionality of the input space increases.

\section{Supplementary results on clustering equivalent expressions}\label{app:equivalent}
This section provides further details on the methodology and results discussed in Section~\ref{sec:clustering}. We generate sets of behaviorally equivalent expressions by starting from a set of expressions grouped according to their functional form:
\begin{itemize}
    \item \textbf{Constant}: $C$
    \item \textbf{Linear}: $C \cdot X_0$, \quad $C \cdot X_1$, \quad $C + C \cdot X_1$, \quad $C + C \cdot X_0 + C \cdot X_1$
    \item \textbf{Quadratic}: $C \cdot X_0^2$, \quad $C + C \cdot X_0 \cdot X_1$, \quad $C \cdot X_0^2 + C \cdot X_1^2$
    \item \textbf{Trigonometric}: $C \cdot \sin(X_0)$, \quad $\cos(C \cdot X_1)$, \quad $C \cdot \sin(X_0) + C \cdot \cos(X_1)$
    \item \textbf{Containing Square Root}: $\sqrt{C \cdot X_0}$, \quad $C + \sqrt{X_0 + X_1}$
    \item \textbf{Logarithmic}: $C \cdot \log(X_0)$, \quad $\log(X_1 + C)$, \quad $C \cdot \log(X_0 \cdot X_1)$
\end{itemize}

Behaviorally equivalent expressions are then produced by applying a series of random algebraic transformations to these initial expressions. The number of transformation steps is drawn uniformly from $1$ to $4$. In each step, the expression tree is traversed, and with a 50\% probability, a randomly chosen mathematically valid transformation is applied to each node. The transformations included:
\begin{itemize}
    \item \textbf{Addition} ($\boldsymbol{+}$): Commutativity ($A+B = B+A$), associativity ($(A+B)+C = A+(B+C)$), factoring ($A+A = 2 \cdot A$), trigonometric identity ($\sin^2 A + \cos^2 A = 1$, where $A$ contains no free parameters), and logarithm combination ($\log A + \log B = \log(A \cdot B)$).
    \item \textbf{Multiplication} ($\boldsymbol{\cdot}$): Commutativity ($A \cdot B = B \cdot A$), associativity ($(A \cdot B) \cdot C = A \cdot (B \cdot C)$), and distributivity ($A \cdot (B+C) = A \cdot B + A \cdot C$).
    \item \textbf{Subtraction} ($\boldsymbol{-}$): Minus removal ($A-B = A + (-1) \cdot B$).
    \item \textbf{Division} ($\boldsymbol{/}$): Division removal ($\frac{A}{B} = A \cdot B^{-1}$).
    \item \textbf{Trigonometric} ($\boldsymbol{\sin/\cos}$): Sine/cosine identity ($\sin A = \cos(A - \frac{\pi}{2})$).
    \item \textbf{Power} ($\boldsymbol{x^p}$): Expansion of a power to a product, applied partially or fully.
    \item \textbf{Zero} ($\boldsymbol{0}$): With a 40\% probability, a $0$ constant is replaced by an equivalent expression such as $\cos(\frac{\pi}{2})$, $\sin(0)$, or $A-A$, where $A$ is a randomly generated expression of length up to 5 without free parameters.
    \item \textbf{One} ($\boldsymbol{1}$): With a 40\% probability, a $1$ constant is replaced by an equivalent expression such as $\sin(\frac{\pi}{2})$, $\cos(0)$, or $A/A$, where $A$ is a randomly generated expression of length up to 5 without free parameters.
\end{itemize}

Additionally, each node has a 4\% probability of receiving an additive identity ($+0$) or a multiplicative identity ($\cdot 1$). This process is repeated until a distinct expression—one not already present in the dataset—is generated.

\begin{table}
\centering
\caption{A comparison of clustering performance using normalized distance matrices as features. Unlike BED, which achieved near-perfect clustering with this approach, syntactic dissimilarity measures (tree edit, edit, and Jaro distance) do not benefit from this representation.}\label{tab:clustering_app}
\vspace{.5cm}
\resizebox{\linewidth}{!}{
\begin{tabular}{lcccc}
Baseline & ARI & Silhouette & V-measure & Fowlkes-Mallows \\
\midrule
Edit distance & $0.002$ ($\pm  0.001$) &  $0.318$ ($\pm  0.051$) &  $0.166$ ($\pm  0.011$) &  $0.217$ ($\pm  0.002$)\\
Edit distance-CN & $0.003$ ($\pm 0.001$) & $0.373$ ($\pm 0.113$) & $0.185$ ($\pm 0.009$) & $0.215$ ($\pm 0.003$)\\
Tree edit distance & $0.004$ ($\pm  0.004$) &  $0.276$ ($\pm  0.060$) &  $0.184$ ($\pm  0.023$) &  $0.218$ ($\pm  0.003$)\\
Tree edit distance-CN & $0.009$ ($\pm 0.006$) & $0.304$ ($\pm 0.075$) & $0.232$ ($\pm 0.037$) & $0.216$ ($\pm 0.005$)\\
Jaro &  $0.002$ ($\pm  0.002$) & $-0.120$ ($\pm  0.140$) &  $0.182$ ($\pm  0.019$) &  $0.213$ ($\pm  0.002$)\\
Jaro-CN &  $0.003$ ($\pm 0.001$) & $-0.372$ ($\pm 0.040$) & $0.182$ ($\pm 0.011$) & $0.216$ ($\pm 0.002$)\\
\end{tabular}}
\end{table}

As discussed in the main text, representing expressions by the rows of the BED distance matrix as feature vectors yields near-perfect clustering results and effectively mitigates the measure’s scale dependence. To test whether this representation could also benefit syntax-based measures, we apply the same procedure to distance matrices obtained using tree edit distance, edit distance, and Jaro distance. As shown in Table~\ref{tab:clustering_app}, this alternative feature representation does not yield any meaningful improvements for these measures. This result reinforces our conclusion that the poor clustering performance of syntax-based measures arises from an intrinsic limitation: their focus on syntactic rather than behavioral dissimilarity.

\begin{figure*}[]
\centering
\includegraphics[width=\linewidth]{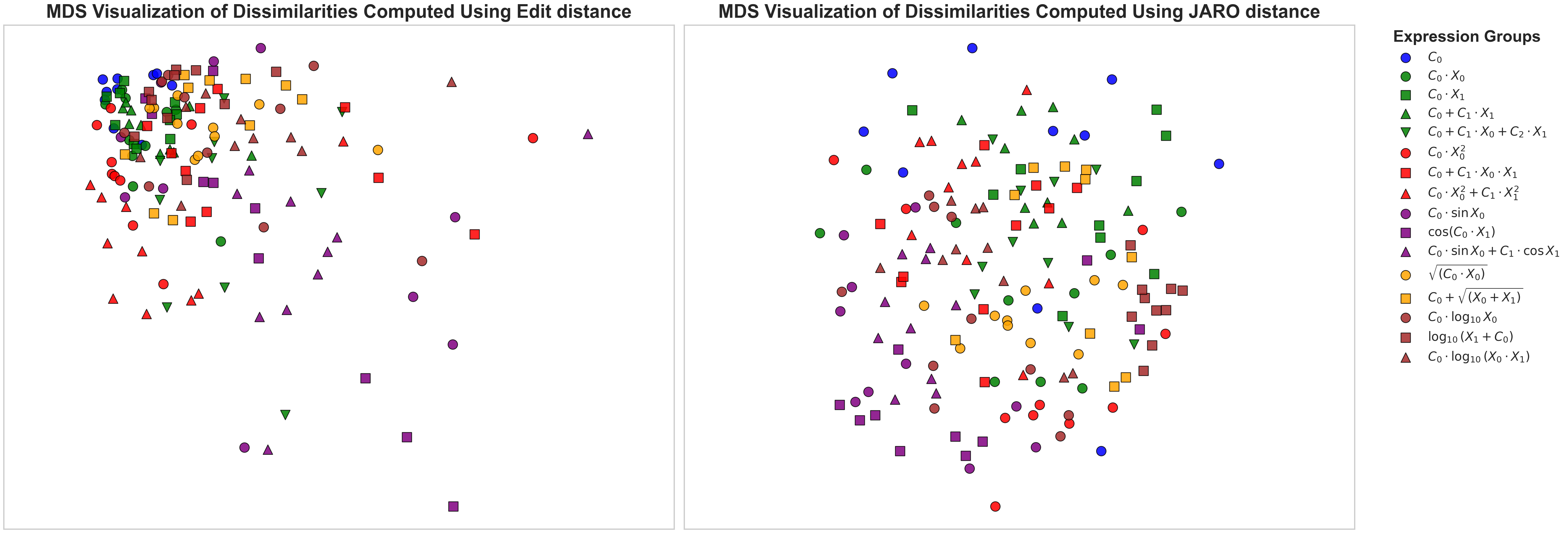}
\caption{Additional MDS visualizations of behavioral clusters using syntactic dissimilarity measures. The plots, showing edit distance (left) and Jaro distance (right), reveal a similar scattered and intermingled distribution to the one seen with tree edit distance. This visual evidence further confirms that these syntactic measures fail to group behaviorally equivalent expressions into meaningful, well-separated clusters.} \label{fig:app_clustering}
\end{figure*}

While Figure~\ref{fig:clustering} in the main text presents MDS visualizations derived from BED and tree edit distance, additional visualizations for the other syntactic measures—edit distance and Jaro distance—are shown in Figure~\ref{fig:app_clustering}. Both plots exhibit highly scattered, overlapping distributions. Behaviorally equivalent expressions, denoted by identical colors and shapes, fail to form the tight, well-separated clusters observed when clustering with BED. These results further emphasize that syntactic measures, regardless of formulation, are poor proxies for behavioral similarity.

\section{Additional results on error landscape smoothness}\label{app:error_manifold}
This section provides supplementary results on the smoothness of the error manifold, complementing the analysis presented in Section~\ref{sec:behavior} of the main text. Here, we repeat the same experiment but aggregate errors using the mean instead of the median. Mean aggregation is more sensitive to outliers, as a few expressions with high errors can disproportionately affect the aggregated result.

\begin{figure*}[htbp]
    \centering
    \begin{subfigure}[b]{0.48\textwidth}
        \centering
        \includegraphics[width=\textwidth]{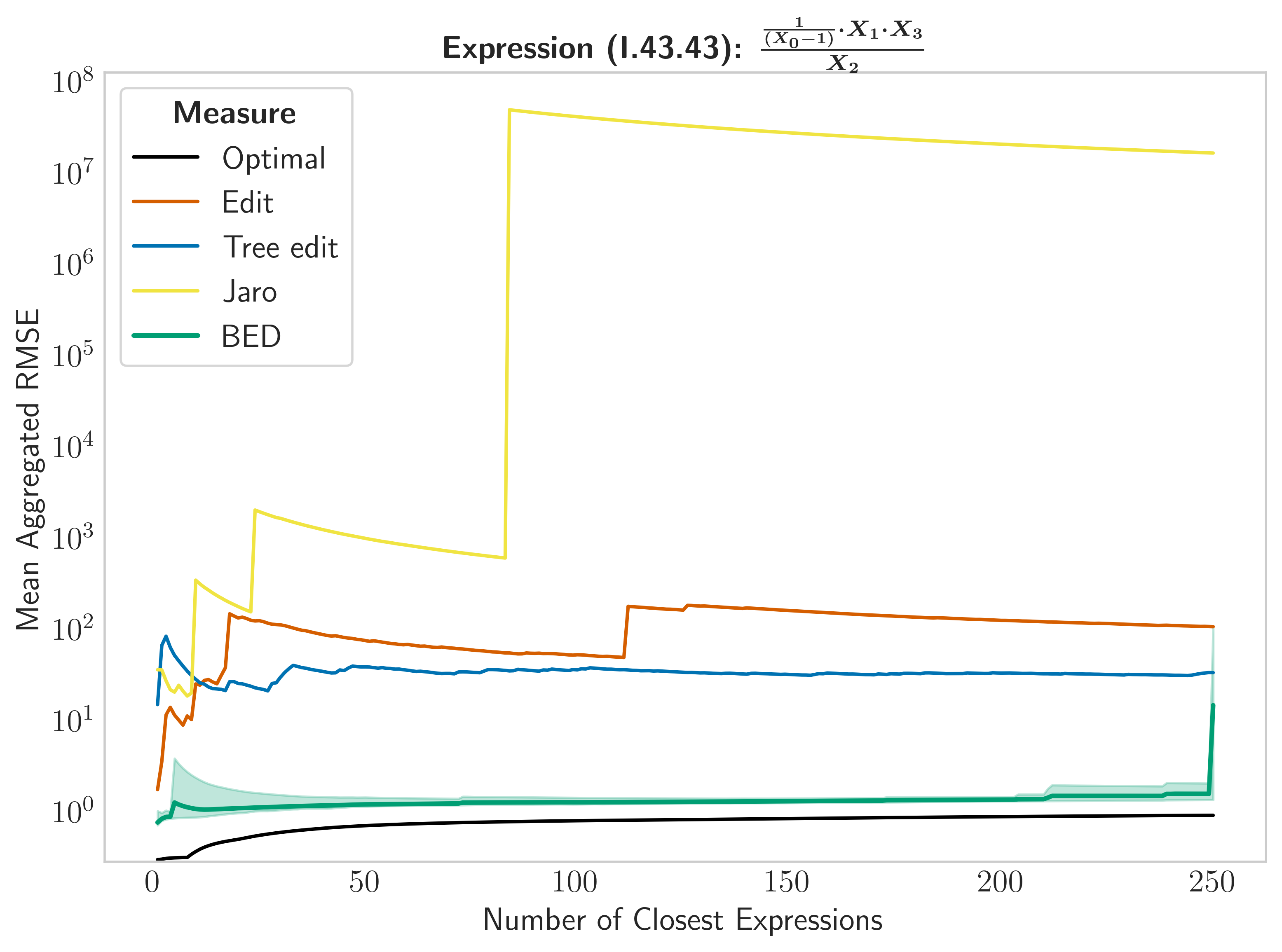} 
        \caption{Expression (I.43.43): $X_0 X_1 X_3 X_2^{-1}$}
        \label{fig:smoothness_aa}
    \end{subfigure}
    \hfill
    \begin{subfigure}[b]{0.48\textwidth}
        \centering
        \includegraphics[width=\textwidth]{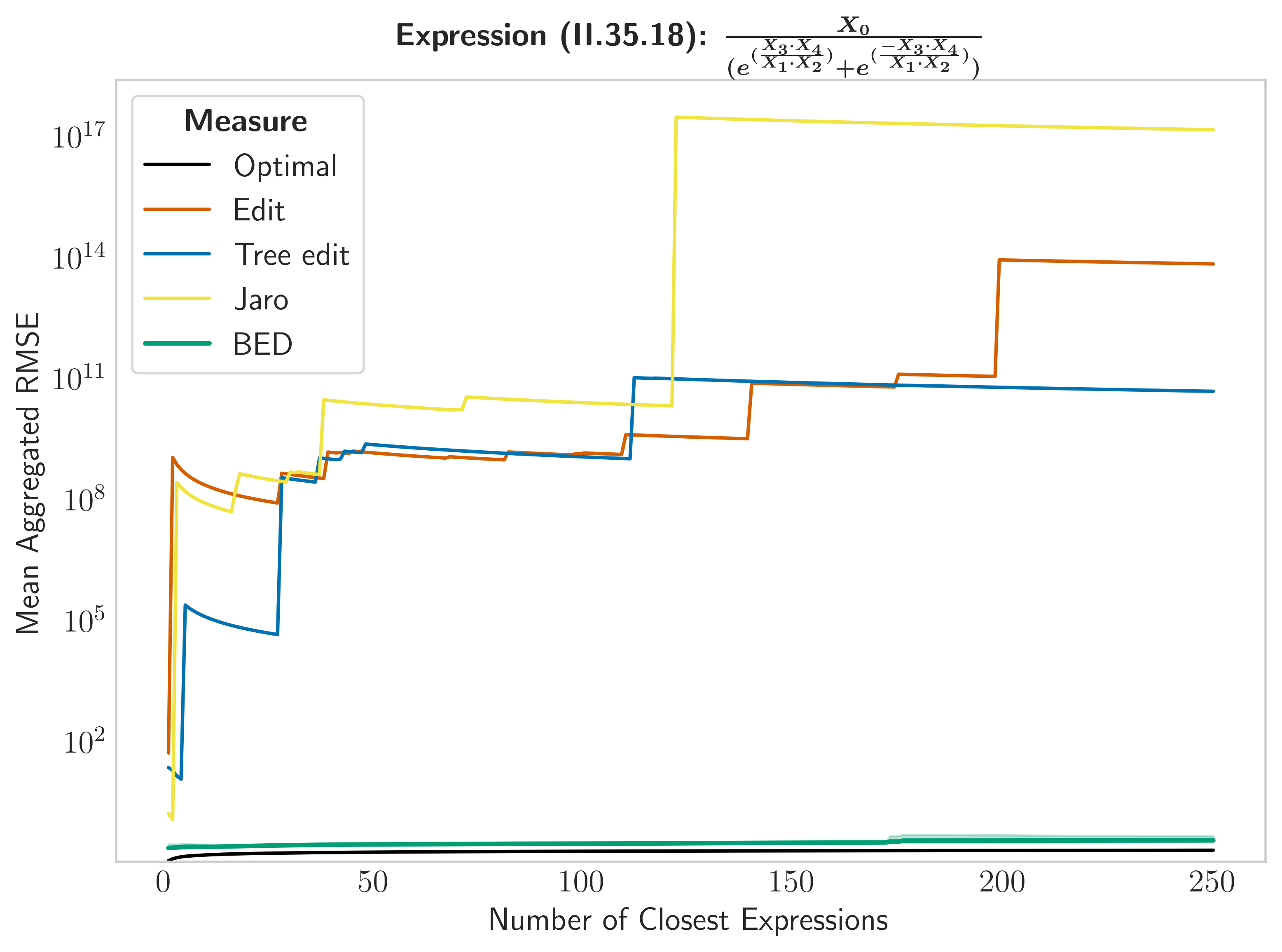} 
        \caption{Expression (II.35.18): $0.5 X_0\sech (\frac{X_3 X_4}{X_1 X_2})$}
        \label{fig:smoothness_ba}
    \end{subfigure}
    \caption{Mean aggregated RMSE for top closest expressions across two benchmark equations. Each subfigure illustrates how the median RMSE of the top N expressions, ordered by different dissimilarity measures, evolves as N increases (from 1 to 250). BED consistently achieves significantly lower and smoother aggregated RMSE curves compared to syntactic measures, demonstrating its superior ability to induce a smooth error manifold.}
    \label{fig:smoothness_app}
\end{figure*}

As shown in Figure~\ref{fig:smoothness_app}, switching from median to mean aggregation does not alter the experimental conclusions much. BED continues to produce a lower and smoother aggregated RMSE curve than all syntactic measures (edit distance, tree edit distance, and Jaro distance). This robustness demonstrates that BED’s capacity to induce a smooth error manifold is an inherent property of the measure itself, rather than an artifact of the chosen aggregation method. Nevertheless, at the far right of the curve in the left panel, even BED occasionally identifies nearest expressions whose errors are substantially higher than those found by the “Optimal” oracle.

\begin{figure}[]
\centering
\includegraphics[width=\linewidth]{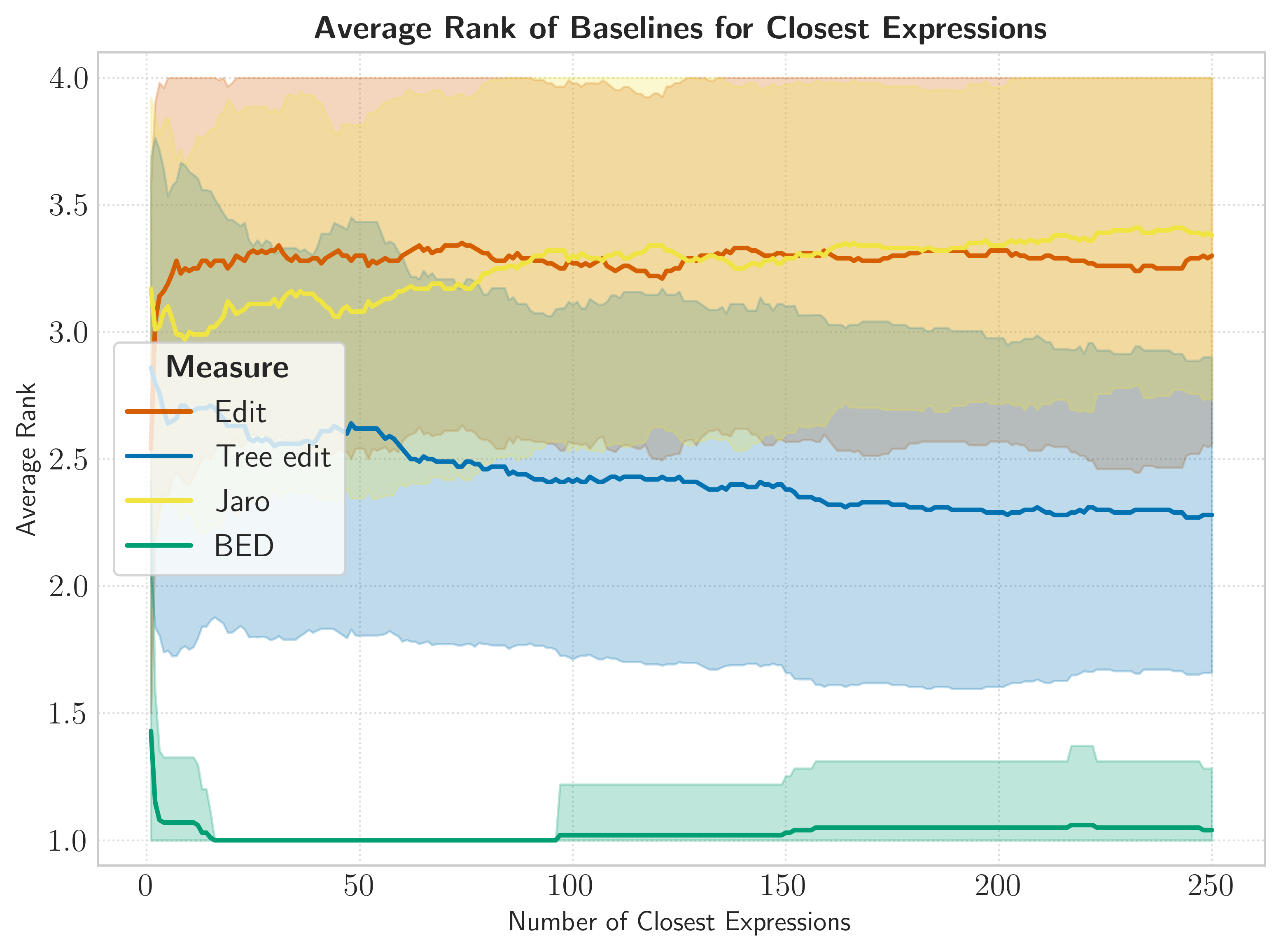}
\caption{Mean-aggregated average rank of dissimilarity measures. The plot shows that BED consistently achieves the lowest average rank (close to 1), even when the mean is used for aggregation. This result, while less stable than the median-based findings, confirms BED's robust superiority over all syntactic baselines.} \label{fig:ranking_app}
\end{figure}

Figure~\ref{fig:ranking_app} presents the aggregated ranks across all $100$ datasets under mean-based aggregation. While BED still achieves an average rank close to $1$, its curve exhibits slightly greater variability and a wider standard deviation compared to the median-based results. This behavior reflects the stronger influence of outliers on mean aggregation. Despite this increased variance, the key finding remains unchanged: BED consistently attains the best average rank and continues to outperform all syntactic baselines.

\bibliographystyle{elsarticle-num-names} 
\bibliography{literature}

\end{document}